\pdfoutput=1
\documentclass[11pt]{article}

\usepackage[final]{acl}

\usepackage{times}
\usepackage{latexsym}

\usepackage[T1]{fontenc}

\usepackage[utf8]{inputenc}

\usepackage{microtype}

\usepackage{inconsolata}

\usepackage{graphicx}

\usepackage{amsmath}
\usepackage{amsfonts}
\usepackage{hyperref}
\usepackage{multirow}
\usepackage{booktabs}
\usepackage{longtable}
\usepackage{caption}
\usepackage{subcaption}
\usepackage{graphicx}
\usepackage{soul}
\usepackage{float}
\usepackage{makecell}
\usepackage{multirow}
\usepackage{listings}

\begin{document}
    \title
{
    ResearchArena: Benchmarking Large Language Models’ Ability to Collect and Organize Information as Research Agents
}

\author
{
    Hao Kang\\
    Carnegie Mellon University\\
    Pittsburgh, PA 15213\\
    \texttt{haok@andrew.cmu.edu}
\And
    Chenyan Xiong\\
    Carnegie Mellon University\\
    Pittsburgh, PA 15213\\
    \texttt{cx@cs.cmu.edu}
}

\maketitle

\begin{abstract}
    Large language models (LLMs) excel across many natural language processing tasks but face challenges in domain-specific, analytical tasks such as conducting research surveys. This study introduces ResearchArena, a benchmark designed to evaluate LLMs' capabilities in conducting academic surveys—a foundational step in academic research. ResearchArena models the process in three stages: (1) information discovery, identifying relevant literature; (2) information selection, evaluating papers' relevance and impact; and (3) information organization, structuring knowledge into hierarchical frameworks such as mind-maps. Notably, mind-map construction is treated as a bonus task, reflecting its supplementary role in survey-writing. To support these evaluations, we construct an offline environment of 12M full-text academic papers and 7.9K survey papers. To ensure ethical compliance, we do not redistribute copyrighted materials; instead, we provide code to construct the environment from the Semantic Scholar Open Research Corpus (S2ORC). Preliminary evaluations reveal that LLM-based approaches underperform compared to simpler keyword-based retrieval methods, though recent reasoning models such as DeepSeek-R1 show slightly better zero-shot performance. These results underscore significant opportunities for advancing LLMs in autonomous research. We open-source the code to construct the ResearchArena benchmark at \url{https://github.com/cxcscmu/ResearchArena}.
\end{abstract}

    \section{Introduction} \label{section:introduction}

Large language models (LLMs) have demonstrated exceptional performance in natural language understanding, text generation, and a range of other tasks across domains \cite{liang2022holistic, bang2023multitask, qin2023chatgpt, laskar2023systematic}. By integrating LLMs with external tools—such as code interpreters, vector databases, and search engines—their capabilities can be further enhanced, enabling the creation of autonomous agents that simulate human-like behavior through feedback-driven task execution \cite{openhands, zhou2023webarena, qin2023toolllm, qian2023communicative}. However, the ability of LLMs to handle domain-specific expertise and advanced analytical tasks, such as conducting rigorous academic research, remains underexplored.

The challenge of conducting domain-specific research is particularly relevant in an era characterized by rapid knowledge expansion across multiple fields. Traditional methods for composing academic surveys are labor-intensive, often requiring months of effort by expert researchers to synthesize relevant findings. An LLM capable of independently conducting research on topics outside its training data could bypass the need for continuous re-training and manual updates, offering a scalable and efficient solution for navigating the ever-growing body of scientific literature.

While autonomous agents have shown success in executing relatively straightforward tasks—such as online shopping or playing card games \cite{zhou2023webarena, liu2023agentbench}—they face far greater challenges in complex tasks that demand extensive domain expertise and analytical depth. Recent developments in agentic capabilities, such as the ``Deep Research'' features from both Gemini and OpenAI, highlight a growing focus on multi-step research planning and the synthesis of large-scale, diverse information sources \cite{gemini-deep-research, openai-deep-research}. However, progress in systematically evaluating these agents' capacity for rigorous research remains limited, with few standardized benchmarks designed for advanced, domain-specific scenarios.

To promote the development of research agents capable of conducting comprehensive surveys, we introduce the ResearchArena benchmark. This benchmark emphasizes academic papers due to their depth of research and structured format, qualities that are often more reliable than other sources like general web pages. The ResearchArena provides an offline environment where autonomous agents can collect and organize information for research across diverse topics. It comprises three sub-tasks for evaluation: information discovery, information selection, and information organization. These sub-tasks emulate general methodologies used by human researchers during literature surveys.

Conducting a literature survey involves defining the scope, establishing a search protocol, and iteratively analyzing and organizing findings into a cohesive structure. Based on this process, ResearchArena introduces tasks to simulate and evaluate these stages, excluding text generation. This decision stems from the premise that the pre-writing research phase is foundational to successful article composition \citep{rohman1965pre}. Moreover, evaluating complete articles is fraught with challenges due to variability in individual writing styles; hence, such assessments are reserved for future work.

For information discovery, LLMs identify and retrieve academic papers relevant to a designated research topic by navigating vast scholarly corpora. Information selection challenges LLMs to critically assess the relevance and impact of these papers, prioritizing significant contributions. As a bonus task, information organization requires LLMs to synthesize selected research into structured knowledge representations, such as mind maps, to highlight key insights and relationships within the topic.

Preliminary evaluations reveal that LLMs underperform compared to simpler keyword-based search methods in tasks requiring analytical depth. For example, using survey titles as retrieval queries consistently yields superior recall and precision compared to LLM-driven information discovery and selection tasks. Additionally, under the task of information organization, LLMs face challenges in constructing coherent structures without the oracle guidance, underscoring the need for improvements in organizational and analytical capabilities. Nonetheless, we observe promising results from reasoning-oriented models such as DeepSeek-R1~\citep{guo2025deepseek}, which show modest gains in zero-shot performance—suggesting that improved reasoning capabilities may help bridge some of these gaps.

The dataset supporting ResearchArena comprises 12M full-text academic papers and 7.9K survey papers, curated from the Semantic Scholar Open Research Corpus (S2ORC) \cite{lo2019s2orc}. This corpus ensures scholarly rigor and relevance, offering a robust foundation for benchmarking LLM performance across diverse domains.

    \section{Related Work} \label{section:related-work}

Previous research has employed diverse methodologies to compile datasets featuring academic survey papers. For instance, BigSurvey dataset \cite{liu2023generating} aggregates over 7K survey papers from arXiv and includes approximately 434K references from Microsoft Academic Service and Semantic Scholar. This dataset underwent extensive preprocessing by removing duplicates, unprocessable files, and normalizing text. On the other hand, Surfer100 dataset \cite{li2021surfer100} includes 100 surveys emulating Wikipedia page structures, compiled by eight annotators who summarized content from web pages. Each survey contains predefined sections such as Introduction, History, Key Ideas, Variations, and Applications, summarized concisely in 50 to 150 words.

The BigSurvey dataset provides references in an abstract-only format, offering a concise overview of documents. Surfer100 utilizes Google search results to compile references for each survey topic, reflecting a broad spectrum of web-based information. In contrast, our dataset emphasizes full-text academic papers for a deeper understanding and leverages bibliographic references from original survey papers for enhanced authority and accuracy.

A closely aligned task for LLM agents in prior research involves generating Wikipedia articles. \citet{liu2018generating} proposed a method for generating English Wikipedia articles by framing the task as a multi-document summarization challenge. Their approach employs a combination of extractive and abstractive summarization techniques, identifying salient information using methods such as TF-IDF and TextRank \cite{mihalcea2004textrank}. Similarly, \citet{shao2024assisting} introduced the STORM system, which tackles pre-writing challenges such as research and outline preparation. STORM enhances the article generation process by simulating multi-perspective conversations, wherein an LLM poses questions and aggregates responses from reliable sources to develop detailed outlines.

Additionally, recent work by Gemini \cite{gemini-deep-research} and OpenAI \cite{openai-deep-research} has explored ``Deep Research'' features that rely on multi-step planning, iterative web browsing, and extended context windows to generate comprehensive reports. These agentic systems aim to automate information discovery and summarization by dynamically adjusting their search strategies and synthesizing insights into structured outputs.

    \section{Collection Methodology} \label{section:collection-methodology}

\begin{figure*}
    \includegraphics[width=\linewidth]{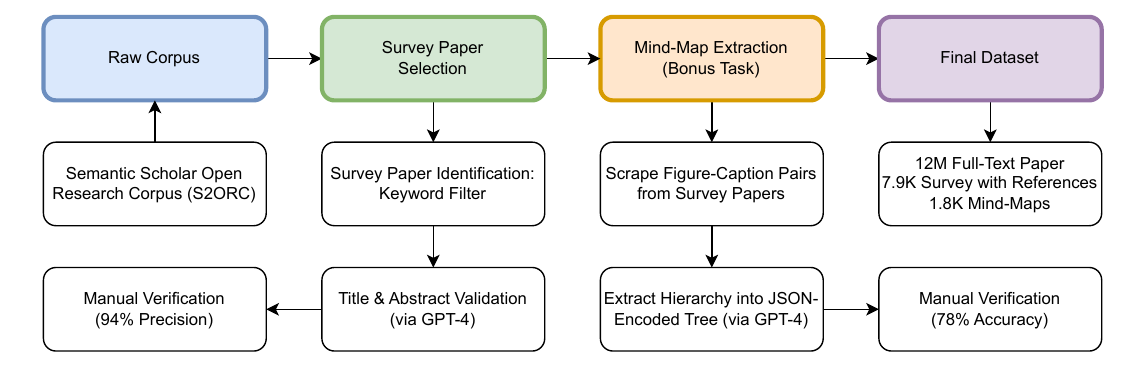}
    \caption{Schematic overview of the construction pipeline for ResearchArena.}
\end{figure*}

\begin{table}[t]
    \small
    \caption{Summary of the dataset composition, including the counts of full-text accessible papers, survey papers, and extracted mind-maps.}
    \centering
    \begin{tabular}{r|l}
        \toprule
        Category & Count \\
        \midrule
        Accessible Papers &  12,034,505\\
        Survey Papers & 7,952\\
        Extracted Mind-Maps & 1,884\\
        \bottomrule
    \end{tabular}
\end{table}

This section describes the multi-stage methodology for assembling the dataset of academic surveys, which includes survey selection, reference linking, and mind-map extraction. Each stage has been designed to ensure the relevance and utility while addressing potential limitations in automation, domain variability, and data access. The final output is a structured dataset that facilitates benchmarking the autonomous research challenge. The details for the prompts used in each stage of the collection process can be found in Appendix \ref{appendix:colleciton-prompts}.

\subsection{Survey Selection}

Survey selection is the foundational step in constructing the dataset, focused on identifying academic papers that provide organized overviews of specific research topics. This process involved leveraging the S2ORC corpus, which contains over 80 million academic articles in machine-readable format. The selection process combined automated filtering and human evaluation to balance scale and accuracy.

Initially, survey papers were identified by filtering for titles containing the term ``survey.'' While this heuristic served as an accessible baseline, it introduced potential biases, such as the exclusion of relevant papers that do not explicitly use the keyword in their titles. For example, in fields like medicine, the terms ``systematic review'' or ``review'' are more common and were largely overlooked. Recognizing these limitations, we further refined our selection using GPT-4 to analyze the titles and abstracts of candidate papers. GPT-4 was prompted to evaluate whether each paper met predefined criteria for surveys, such as presenting a comprehensive overview of a field.

Through this two-stage approach, approximately 85\% of papers initially flagged by keyword filtering were excluded after GPT-4 evaluation. To validate this methodology, we conducted a manual inspection of a random sample of 100 papers from the final collection, achieving a 94\% precision in identifying relevant surveys. The details of this inspection are provided in Appendix \ref{appendix:inspection-over-survey-passages}. Although this method cannot guarantee perfect recall, we believe it sufficiently represents the broader distribution of survey literature in various domains. Additionally, this stage prioritized full-text accessibility within S2ORC to ensure the inclusion of rich contextual details, reducing the corpus size to approximately 12 million documents.

\subsection{Reference Linking}

The second stage of the methodology involved extracting and linking bibliographic references cited in the identified survey papers. This step is critical for evaluating tasks related to information discovery and selection, as it connects surveys to their foundational sources. Reference data were sourced directly from the S2ORC corpus, which includes pre-resolved bibliographic metadata.

Despite the robustness of S2ORC’s reference extraction capabilities, several challenges emerged, including missing references or misclassified citation structures. Surveys without detectable bibliographic sections—often due to formatting issues in the source data—were excluded, resulting in the removal of 406 survey papers. Additionally, 1,635 surveys were discarded because they lacked accessible references, rendering them unsuitable for downstream evaluations.

It is important to acknowledge that the reference linking process—like human citation practices—is inherently imperfect. Even expert researchers may unintentionally omit relevant works or introduce redundancies. Similarly, our automated approach provides an approximation that, while robust, does not guarantee perfect recall of influential references. To mitigate this limitation, we applied a supervised classification model inspired by \citet{valenzuela2015identifying} to distinguish influential from non-influential citations, ensuring that the most impactful references were prioritized.

Moreover, publication dates were annotated for each retained reference, with conservative imputation for missing months or days to minimize information leakage in temporal evaluations. While these efforts improve the utility and reliability of the dataset, we recognize that no methodology can fully account for all relevant literature. Future enhancements, including integrating domain-specific heuristics and engaging human annotators, may further refine this process.

\subsection{Mind-Map Extraction}

Mind-map extraction is positioned as a bonus task within the benchmark, complementing the primary objectives of survey selection and reference linking. While mind-maps are not commonly found in academic survey papers, they provide valuable hierarchical visualizations of knowledge when present, offering an organized perspective on the topics.

\begin{figure*}
\begin{minipage}[b]{0.42\linewidth}
\includegraphics[width=\linewidth]{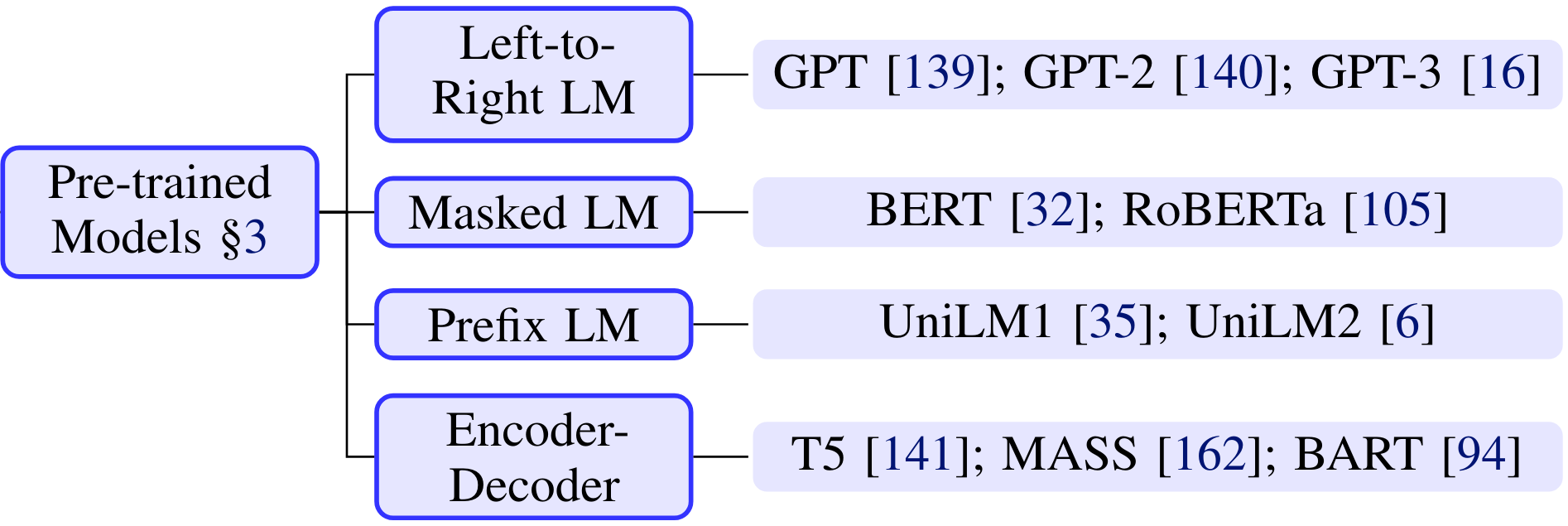}
\end{minipage}
\hfill
\begin{minipage}[b]{0.48\linewidth}
\fontsize{8}{8}
\begin{verbatim}
{
    "Pre-trained Models": {
        "Left-to-Right LM": ["GPT", "GPT-2", "GPT-3"],
        "Masked LM": ["BERT", "RoBERTa"],
        "Prefix LM": ["UniLM1", "UniLM2"],
        "Encoder-Decoder": ["T5", "MASS", "BART"],
    }
}
\end{verbatim}
\end{minipage}
\caption{Mind-map extraction from the figure to its JSON representation.}
\label{figure:mind-map-extraction}
\end{figure*}

Limited by the text-only nature of the S2ORC corpus, we extended our dataset by sourcing figure-caption pairs from the Semantic Scholar website, specifically targeting surveys with accessible figures. Using GPT-4, figures and their captions were analyzed to identify those likely representing mind-maps. Relevant figures were converted into JSON-encoded hierarchical structures, preserving their organizational logic, as illustrated in Figure \ref{figure:mind-map-extraction}.

This task employed a two-step verification process: first, determining if the figure represented a valid taxonomy, and second, assessing its relevance to the survey topic. After the extraction, a manual review of 100 mind-maps yielded an accuracy rate of 78\% for hierarchical representation and 70\% for topic relevance. The details of this inspection are provided in Appendix \ref{appendix:inspection-over-survey-passages}. While these scores highlight the limitations of automated extraction and domain variability, they underscore the utility of mind-maps as an auxiliary dataset feature for future exploratory research.

\subsection{Dataset Access}

To ensure compliance, we provide tools and code that enable users to reproduce the dataset using publicly accessible S2ORC. This approach avoids direct distribution of the corpus while empowering researchers to generate reproducible datasets tailored to their specific needs.

Users must independently verify licensing requirements for the underlying data sources, as open access does not inherently guarantee permissive redistribution rights. Using February 06, 2024 release of S2ORC, the dataset itself consists of approximately 12 million full-text academic papers, including 7,952 survey papers and 1,884 extracted mind-maps.

Most importantly, our released pipeline supports weekly updates from Semantic Scholar, enabling evaluations to continuously incorporate recent advancements and maintain ongoing relevance.

    \section{Analysis} \label{section:dataset_composition}

This section details the makeup of our dataset in terms of disciplinary diversity, reference coverage, and the structural complexity of derived typologies, reflecting on how these factors contribute to the robustness and applicability across various domains.

\begin{figure*}
    \begin{subfigure}[b]{0.23\linewidth}
        \centering
        \includegraphics[width=\linewidth]{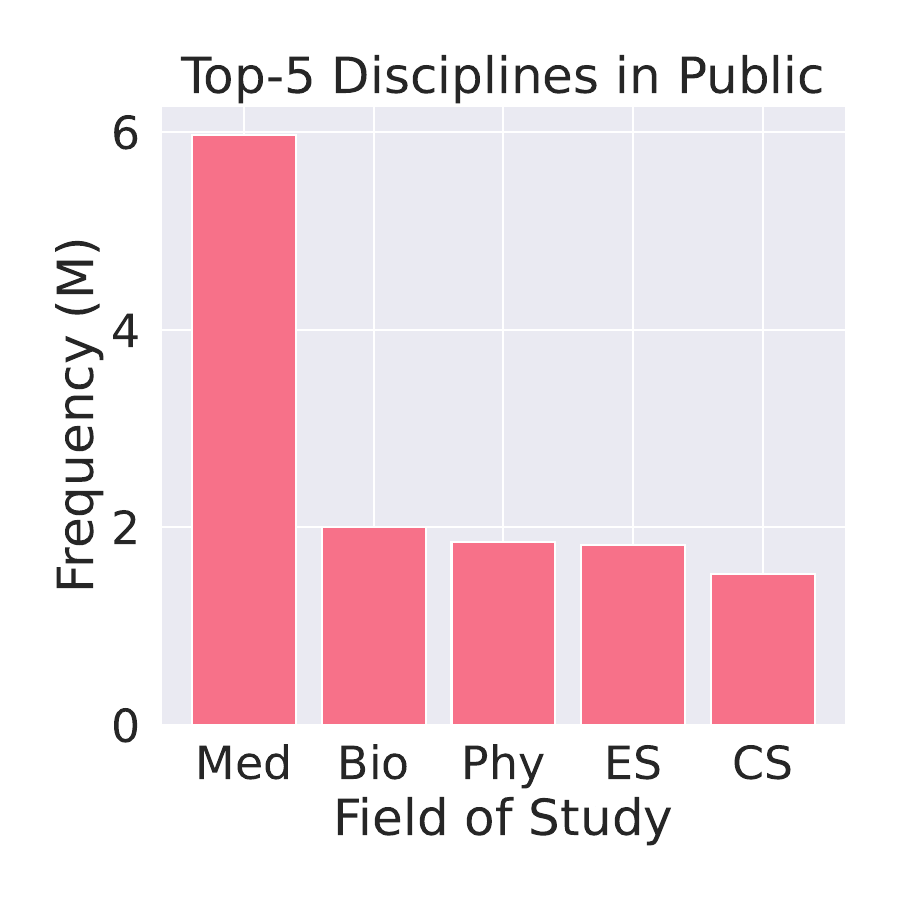}
        \subcaption{Disciplinary distribution of the public corpus.}
        \label{figure:public_discipinary}
    \end{subfigure}
    \hspace{0.01\linewidth}
    \begin{subfigure}[b]{0.23\linewidth}
        \centering
        \includegraphics[width=\linewidth]{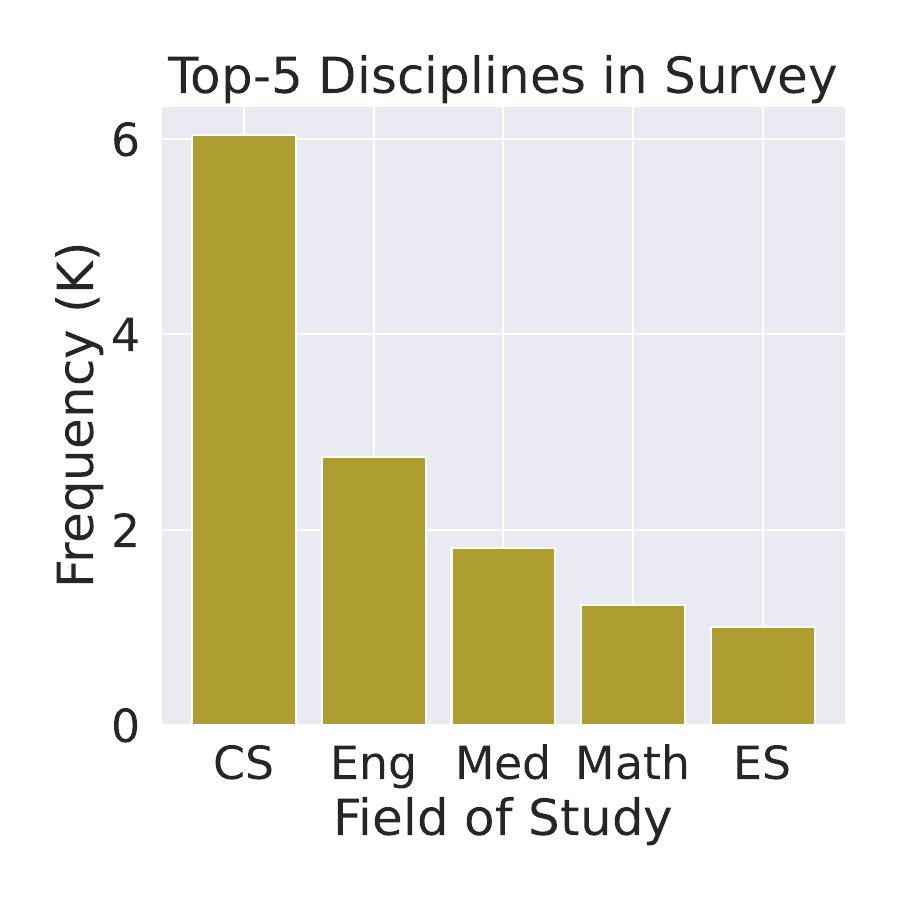}
        \subcaption{Disciplinary distribution of the survey subset.}
        \label{figure:survey_discipinary}
    \end{subfigure}
    \hspace{0.01\linewidth}
    \begin{subfigure}[b]{0.23\linewidth}
        \centering
        \includegraphics[width=\linewidth]{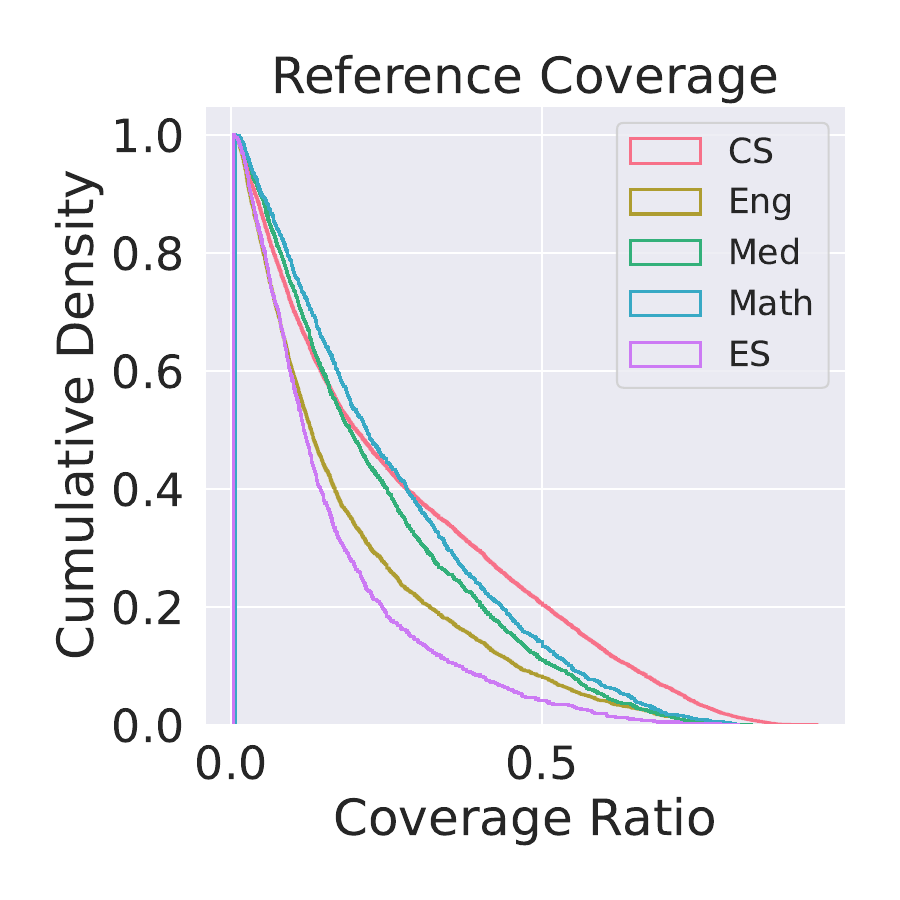}
        \subcaption{Reference coverage of the survey subset.}
        \label{figure:reference_coverage}
    \end{subfigure}
    \hspace{0.01\linewidth}
    \begin{subfigure}[b]{0.23\linewidth}
        \centering
        \includegraphics[width=\linewidth]{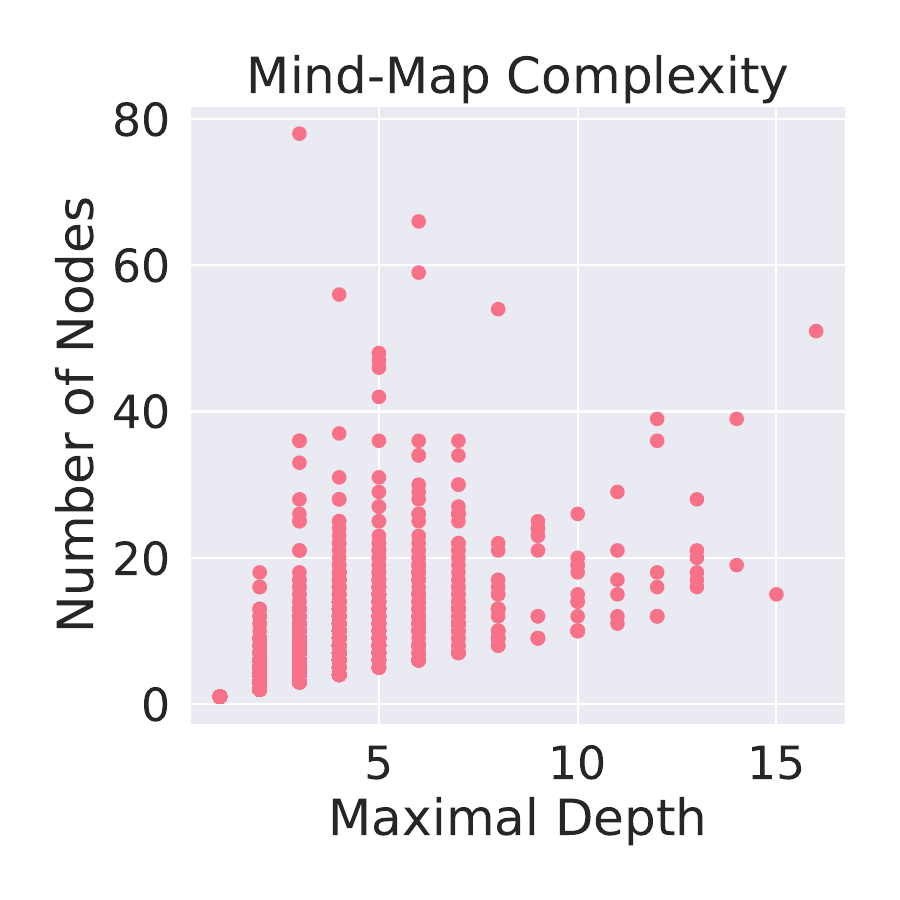}
        \subcaption{Complexity with the extracted mind-maps.}
        \label{figure:typology_complexity}
    \end{subfigure}
    \caption{Dataset composition analysis with disciplinary distribution, reference coverage, and mind-map complexity. Each of these aspects is critical for benchmark evaluation. Fields of studies like Medicine (Med), Biology (Bio), Physics (Phy), Environmental Science (ES), Computer Science (CS), Engineering (Eng), and Mathematics (Math) are denoted with their abbreviations in the figures.}
\end{figure*}

\textbf{Disciplinary Distribution.} We classified each of the 12.0M papers in our public corpus and 7.9K survey papers by the top-5 most popular academic disciplines. This classification was based on the indexing information provided by S2ORC. Frequencies of papers per discipline were then aggregated and visualized to identify trends and imbalances.  Figure \ref{figure:public_discipinary} and \ref{figure:survey_discipinary} revealed significant disparities in the frequency of disciplines between the public corpus and the survey subset. Notably, Computer Science is the most prevalent discipline within surveys but less common in the broader corpus. This could reflect the dynamic nature of the CS field, which often necessitates comprehensive reviews to synthesize rapid advancements and emerging trends. On the other hand, the dominance of medical papers in the public corpus stems from the composition of the original S2ORC dataset, which was itself heavily sourced from biomedical publications.

\textbf{Reference Coverage.} For each survey paper, we calculated the coverage ratio as the proportion of its references that were also available within our full-text corpus. We plotted cumulative density functions for each discipline to analyze how extensively the surveys' references are represented in the broader corpus. As illustrated with Figure \ref{figure:reference_coverage}, similar patterns were observed across all disciplines, where the density experienced exponential decay as the coverage increases. Approximately 17.18\% of the survey subset (i.e., 1.3K survey papers) have at least 50\% of their references available. This limitation is mainly attributed to copyright restrictions, where full-text is not permitted by the publisher.

\textbf{Mind-Map Complexity.} We analyzed the structural complexity of the mind-maps extracted from survey papers by counting the number of nodes and measuring the maximal depth. These measures provide insights into the conceptual breadth and hierarchical depth of the topics covered. The scatter plot from Figure \ref{figure:typology_complexity} showed that typologies in general have shallow depths but a broad range of nodes, suggesting that while survey topics are extensively branched, they do not delve deeply into sub-topics. In particular, most typologies have a maximum depth ranging from 3 to 7 levels, where the coefficient of the regression line in the scatter plot is approximately 2.04.

    \section{Benchmark Tasks} \label{section:benchmark_tasks}

This section presents a comprehensive overview of the benchmark tasks designed to evaluate the capabilities of research agents in discovering, selecting, and organizing information. Each task targets a specific aspect of research proficiency, with rigorous constraints and evaluation metrics to ensure thorough and unbiased assessment. 

\textbf{Information Discovery.}
Provided a topic extracted from survey title, the task of information discovery requires research agents to identify a subset of documents $R$ from a broader collection $D$. These documents in $R$ should serve as supporting materials for the topic. Ideally, $R$ should encompass all references cited in the original survey $S$.

However, within the collection $D$, there may exist another survey $S'$ that delves into the same topic. If research agents were to use the references from $S'$ directly, it would circumvent the need for a thorough discovery, defeating the purpose of this task. To prevent information leakage, we impose the additional constraint such that documents in $D$ must be non-survey and published before $S$. 

To evaluate performance, we employ standard information retrieval metrics, Recall and Precision, to measure the proportion of relevant documents successfully retrieved and the proportion of retrieved documents that are relevant. Together, these metrics determine the effectiveness and accuracy of the discovery process. For this task, the cutoff parameter $K$ is set at 10 and 100.

\textbf{Information Selection.} The task of information selection requires research agents to rank the discovered documents based on their importance to the topic. The labels are distinctions between influential and non-influential citations, as elaborated in Section \ref{section:collection-methodology}. Normalized Discounted Cumulative Gain (nDCG) \cite{jarvelin2002cumulated} and Mean Reciprocal Rank (MRR) \cite{voorhees1999proceedings} are used for evaluation.

These measures are crucial because conducting research involves more than merely summarizing retrieved documents; it requires the presentation of key insights from the most significant sources. Furthermore, both human researchers and autonomous agents are limited by their processing capacities. Therefore, it is essential to prioritize and focus on the most critical information first.

\textbf{Information Organization (Bonus).} For information organization, research agents are required to construct a hierarchical knowledge mind-map $M$ based on $R$. This mind-map should provide a systematic overview of research work developed on topic $T$. As an intermediate step, references $R$ from the original survey paper could be provided to the agents, who would then focus exclusively on constructing $M$. In contrast, for an end-to-end version, $R$ is the set of discovered documents from the previous task.

For evaluation, two primary metrics are employed: Heading Soft Recall \cite{FRANTI2023115} and Heading Entity Recall \cite{shao2024assisting}. These metrics compare the set of node labels from the original and the constructed knowledge mind-maps, referred to as $A$ and $B$, respectively. To measure similarity of these labels, Heading Soft Recall (HSR) leverages \textsc{Sentence-Bert} \cite{reimers-gurevych-2019-sentence} embedding, while Heading Entity Recall (HER) employs Named Entity Recognition from FLAIR \cite{akbik-etal-2019-flair} for extraction.

While these metrics provide a measure of content similarity, they do not account for structural alignment. Tree Editing Distance \cite{zhang1989simple} solves this concern by calculating the minimal number of operations (i.e., relabeling, deleting, and inserting nodes) required to transform one tree into another. Nonetheless, relying on Tree Editing Distance alone might overlook the potential for non-exact label matches. To address this, we propose Tree Semantic Distance, which assigns no cost to editing operations involving nodes whose cosine similarity exceeds $0.8$.

    \section{Benchmarking} \label{section:experiments}

\begin{table*}[t]
    \small
    \caption{Baseline performance on discovery task, evaluated with Recall@10, Recall@100, Precision@10, and Precision@100, where the retrievers include BM25 and BGE.}
    \label{table:discovery_performance}
    \centering
    \begin{tabular}{rcccccccc}
        \toprule
        & \multicolumn{2}{c}{Recall@10} & \multicolumn{2}{c}{Recall@100} & \multicolumn{2}{c}{Precision@10} & \multicolumn{2}{c}{Precision@100} \\
        \cmidrule(lr){2-3} \cmidrule(lr){4-5} \cmidrule(lr){6-7} \cmidrule(lr){8-9}
         Baseline  & BM25 & BGE & BM25 & BGE & BM25 & BGE & BM25 & BGE \\
        \midrule
        \textsc{Title} & 0.0424 & \textbf{0.1012} & 0.1338 & \textbf{0.2697} & 0.0669 & \textbf{0.1541} & 0.0286 & \textbf{0.0586} \\
        \textsc{Zero-Shot (GPT-4)} & 0.0382 & 0.0832 & 0.1253 & 0.2287 & 0.0602 & 0.1232 & 0.0256 & 0.0464 \\
        \textsc{Decomposer (GPT-4)} & 0.0434 & 0.0879 & 0.1431 & 0.2554 & 0.0717 & 0.1304 & 0.0312 & 0.0536 \\
        \textsc{Zero-Shot (Claude 3.5 Sonnet)} & 0.0336 & 0.0777 & 0.1173 & 0.2169 & 0.0537 & 0.1137 & 0.0235 & 0.0428 \\
        \textsc{Decomposer (Claude 3.5 Sonnet)} & 0.0435 & 0.0876 & 0.1496 & 0.2547 & 0.0751 & 0.1290 & 0.0331 & 0.0541 \\
        \textsc{Zero-Shot (DeepSeek-R1)} & 0.0446 & 0.0865 & 0.1491 & 0.2459 & 0.0730 & 0.1304 & 0.0316 & 0.0511 \\
        \textsc{Decomposer (DeekSeek-R1)} & 0.0418 & 0.0858 & 0.1478 & 0.2514 & 0.0731 & 0.1281 & 0.0331 & 0.0541 \\
        \textsc{Self-RAG} & 0.0380 & 0.0815 & 0.1210 & 0.2260 & 0.0595 & 0.1215 & 0.0256 & 0.0461 \\
        \textsc{STORM} & 0.0281 & 0.0979 & 0.0693 & 0.1441 & 0.0446 & 0.1041 & 0.0130 & 0.0208 \\
        \bottomrule
    \end{tabular}
\end{table*}

\begin{table*}[t]
    \small
    \caption{Baseline performance on selection task, evaluated with nDCG@10, nDCG@30, nDCG@100, and MRR, where the retrievers include BM25 and BGE.}
    \label{table:selection_performance}
    \centering
    \begin{tabular}{rcccccccc}
        \toprule
        & \multicolumn{2}{c}{nDCG@10} & \multicolumn{2}{c}{nDCG@30} & \multicolumn{2}{c}{nDCG@100} & \multicolumn{2}{c}{MRR} \\
        \cmidrule(lr){2-3} \cmidrule(lr){4-5} \cmidrule(lr){6-7} \cmidrule(lr){8-9}
         Baseline  & BM25 & BGE & BM25 & BGE & BM25 & BGE & BM25 & BGE \\
        \midrule
        \textsc{Title} & 0.0711 & \textbf{0.1678} & 0.0775 & \textbf{0.1754} & 0.0941 & \textbf{0.2019} & 0.1903 & \textbf{0.3816} \\
        \textsc{Zero-Shot (GPT-4)} & 0.0634 & 0.1346 & 0.0692 & 0.1417 & 0.0856 & 0.1657 & 0.1743 & 0.3246 \\
        \textsc{Decomposer (GPT-4)} & 0.0735 & 0.1445 & 0.0803 & 0.1554 & 0.0986 & 0.1838 & 0.1959 & 0.3510 \\
        \textsc{Zero-Shot (Claude 3.5 Sonnet)} & 0.0584	& 0.1258 & 0.0630 & 0.1326 & 0.0792 & 0.1562 & 0.1763 & 0.3097 \\
        \textsc{Decomposer (Claude 3.5 Sonnet)} & 0.0760 & 0.1437 & 0.0827 & 0.1548 & 0.1025 & 0.1834 & 0.2015 & 0.3518 \\
        \textsc{Zero-Shot (DeepSeek-R1)} & 0.0756 & 0.1401 & 0.0824 & 0.1488 & 0.1016 & 0.1763 & 0.2081 & 0.3336 \\
        \textsc{Decomposer (DeekSeek-R1)} & 0.0740 & 0.1424 & 0.0811 & 0.1530 & 0.1009 & 0.1817 & 0.1977 & 0.3520 \\
        \textsc{Self-RAG} & 0.0627 & 0.1341 & 0.0679 & 0.1415 & 0.0837 & 0.1646 & 0.1705 & 0.3233 \\
        \textsc{STORM} & 0.0445 & 0.1275 & 0.0507 & 0.1322 & 0.0524 & 0.1267 & 0.1271 & 0.3206 \\
        \bottomrule
    \end{tabular}
\end{table*}

\begin{table*}[t]
    \small
    \caption{Baseline performance on organization task, evaluated with Heading Soft Recall, Heading Entity Recall, and Tree Semantic Distance, across intermediate and end-to-end conditions.}
    \centering
    \begin{tabular}{crccc}
        \toprule
            Oracle & Baseline & Heading Soft Recall ($\uparrow$) & Heading Entity Recall ($\uparrow$) & Tree Semantic Distance ($\downarrow$) \\
        \midrule
        \multirow{2}{*}{Yes}
        & \textsc{Clustering} & 0.6074 & 0.2104 & \textbf{45.69} \\
        & \textsc{STORM} & \textbf{0.7325} & \textbf{0.3098} & 60.04 \\
        \midrule
        \multirow{3}{*}{No}
        & \textsc{Few-Shot} & \textbf{0.8408} & 0.2446 & \textbf{49.83} \\
        & \textsc{STORM.BM25} & 0.7940 & \textbf{0.2938} & 66.65 \\
        & \textsc{STORM.BGE} & 0.7842 & 0.2693 & 65.93 \\
        \bottomrule
    \end{tabular}
    \label{table:organization_performance}
\end{table*}

In this section, we present preliminary evaluations of existing techniques, describing their configurations and performance metrics. These techniques encompass both naive keyword-based methods, such as \textsc{Title}, and advanced LLM-based methods, including \textsc{STORM}. The exact wording of the prompts used in each baseline can be found in Appendix \ref{appendix:experiment-prompts}. We adopt the abstract as the indexing scheme, as it consistently yields the best performance. Results for alternative indexing schemes are reported in Appendix~\ref{appendix:indexing-schemes}.

\subsection{Baselines}

\textbf{Information Discovery.} For information discovery, research agents are equipped with retrieval tools that enable interaction with the public corpus by submitting queries to retrievers such as BM25 and BGE \cite{xiao2023c}. These agents are evaluated based on their ability to effectively leverage these tools by generating relevant queries. Since exploration is limited to previously published non-survey literature, retrievers retry with exponential back-off until the cutoff parameter $K$ is satisfied.

\textsc{Title}: This method assumes that survey titles encapsulate their respective research topics. It directly uses the title from each survey paper as a query to retrieve relevant materials. Since titles in the S2ORC corpus exhibit inconsistent capitalization, we normalize them by converting all titles to lowercase.

\textsc{Zero-Shot}: Building on the \textsc{Title} method, this approach leverages LLMs—such as GPT-4, Claude 3.5 Sonnet, or DeepSeek-R1—to generate refined queries based on the survey title. The assumption is that these models possess prior knowledge relevant to the topic and can produce more precise and context-aware queries.

\textsc{Decomposer}: Based on the findings of Tushar et al. \cite{khot2022decomposed}, decomposed prompting is beneficial when complex reasoning tasks are difficult to perform in a single step. Survey topics often consist of multiple sub-topics, making them well-suited for this approach. We instruct LLMs to break the topic into several sub-questions, each of which is used to generate a sub-query. These sub-queries are executed in batches, and the retrieved results are merged using reciprocal rank fusion \cite{cormack2009reciprocal}.

\textsc{Self-RAG}: Proposed by Asai et al. \cite{asai2023self}, \textsc{Self-RAG} retrieves passages on demand and uses reflection tokens to assess their relevance. It then incorporates relevant information into the ongoing generation. This approach improves upon \textsc{Zero-Shot} by allowing the model to iteratively refine its queries based on intermediate retrievals, effectively acting as a research agent.

\textsc{STORM}: As described in Section \ref{section:related-work}, \textsc{STORM} uses multi-perspective conversations to construct Wikipedia-style articles from scratch. It mirrors our use case, except that our context involves academic literature rather than encyclopedic content. We log all retrieved documents during the conversation process, and the final retrieval set is treated as the body of discovered evidence.

\textsc{Deep Research}: This system showcases advanced multi-step browsing and iterative information synthesis via web-based exploration. However, it currently does not support offline corpora and relies on open web resources. We include a brief case study to illustrate its potential and limitations.

\textbf{Information Selection.} For information selection, documents are ranked based on the similarity scores obtained during the discovery phase. For BGE retriever, we rely on FAISS \cite{johnson2019billion} to retrieve based on L2 distance in the embedding space. On the other hand, \textsc{STORM} does not explicitly rank the retrieved documents. We treat documents discovered earlier in the conversations of higher relevance.

\textbf{Information Organization.} For information organization, the \textsc{Clustering} approach employs Ward's method for hierarchical clustering on the BGE embedding of every reference article, and the final dendrogram is extracted as typology. The label in each node is computed as the most important TF-IDF word, with ngrams ranging from 1 to 3. \textsc{Few-Shot} is achieved by providing a few random examples of extracted typologies and instructing GPT-4 to generate another topic-oriented mind-map. Lastly, the article outline generated by \textsc{STORM} is converted to typology, with headings and their nested sub-headings representing the hierarchy.

\subsection{Evaluation Results}

The baseline experiments were conducted on a single machine equipped with 8 NVIDIA RTX A6000 GPUs, 96 CPU cores, and 128GB RAM. Discussion on the performance metrics is presented below.

\textbf{Information Discovery.} As demonstrated in Table \ref{table:discovery_performance}, the task of information discovery remains challenging for all baseline models. This is illustrated by the Recall@100 metric, which falls below 0.15 for BM25 and 0.27 for BGE. Moreover, agent baselines such as \textsc{Self-RAG} and \textsc{STORM} consistently achieve the lowest rankings, irrespective of the retrievers employed. This limitation highlights the critical need for more advanced retrieval mechanisms to manage large volumes of documents effectively during information discovery.

\textbf{Information Selection.} The performance with information selection is presented in Table \ref{table:selection_performance}. The results indicate a consistent trend wherein agent baselines underperform compared to keyword-based methods. The evaluation of nDCG at various levels of document retrieval, such as nDCG@10, nDCG@30, and nDCG@100, provides a quantitative assessment of the ranking performance. Notably, for the \textsc{Title} method using the BGE retriever, the nDCG@100 score is 0.2019, which significantly surpasses the score of \textsc{STORM}, which stands at 0.1267. Improvements during the information discovery phase have the potential to enhance overall performance in the selection phase, as evidenced by \textsc{Decomposer}, which ranks the second behind \textsc{Title} in discovery and selection tasks.

\textbf{Information Organization.} The evaluation on task of information organization under intermediate (i.e., with oracle) and end-to-end (i.e., without oracle) conditions are documented in Table \ref{table:organization_performance}. Notably, the metrics exhibit discrepancies across each other, which contrasts with the uniformity observed in previous discovery and selection tasks. This divergence is expected due to the distinct nature of the metrics: Heading Soft Recall and Heading Entity Recall assess content similarity, whereas Tree Semantic Distance evaluates structural alignment.

In the intermediate version, where references are provided to LLMs, the proportion of correctly included entities, as measured by Heading Entity Recall, is slightly higher. Specifically, \textsc{STORM} achieved a recall rate of 0.3098, outperforming the end-to-end condition. Conversely, when it comes to constructing the hierarchy, \textsc{Clustering} outperforms advanced LLM-based agents, as evidenced by its attainment of the lowest Tree Semantic Distance of 45.69 among all baseline methods.

\textbf{Deep Research.} We conducted a brief investigation on two distinct topics, transfer learning and LiDAR scanning mechanisms, using Gemini Deep Research, as shown in Appendix \ref{appendix:case-study-with-gemini-deep-research}. Due to its sole reliance with online resources, we don't have a direct quantitative comparison with other baselines. Nevertheless, the generated summaries illustrate Gemini Deep Research's capacity to synthesize diverse online sources into coherent findings, highlighting its potential to support high-level exploration when specialized academic databases are not immediately required.

    \section{Conclusion}

In conclusion, ResearchArena introduces a rigorous benchmark designed to evaluate LLMs in conducting research surveys on designated topics. By systematically decomposing the survey process into distinct tasks like information discovery, selection, and organization, this benchmark provides a detailed framework for evaluating autonomus research agents. Our findings underscore the potential of LLMs to revolutionize academic research, provided that future advancements can bridge the existing performance gaps. Grounded in Semantic Scholar Open Research Corpus, this work establishes a robust foundation for the future, aiming to improve the ability of LLMs to autonomously conduct expertise-level, domain-specific research.

    \section{Limitations}

Despite the robust framework and extensive dataset provided by ResearchArena, this study has several limitations. Firstly, due to copyright constraints, not every full-text reference of the survey papers could be included. This omission could affect the comprehensive understanding of the survey topics under investigation. In addition, there is no evaluation on text generation but mostly the surveying process. However, even if this is just the first step of conducting research, LLM agents have already shown deficiencies. Future iterations of ResearchArena should address this issue, particularly as these agents improve.

    \bibliography{sections/References}

\begin{thebibliography}{32}
\providecommand{\natexlab}[1]{#1}

\bibitem[{Akbik et~al.(2019)Akbik, Bergmann, Blythe, Rasul, Schweter, and Vollgraf}]{akbik-etal-2019-flair}
Alan Akbik, Tanja Bergmann, Duncan Blythe, Kashif Rasul, Stefan Schweter, and Roland Vollgraf. 2019.
\newblock \href {https://doi.org/10.18653/v1/N19-4010} {{FLAIR}: An easy-to-use framework for state-of-the-art {NLP}}.
\newblock In \emph{Proceedings of the 2019 Conference of the North {A}merican Chapter of the Association for Computational Linguistics (Demonstrations)}, pages 54--59, Minneapolis, Minnesota. Association for Computational Linguistics.

\bibitem[{Asai et~al.(2023)Asai, Wu, Wang, Sil, and Hajishirzi}]{asai2023self}
Akari Asai, Zeqiu Wu, Yizhong Wang, Avirup Sil, and Hannaneh Hajishirzi. 2023.
\newblock Self-rag: Learning to retrieve, generate, and critique through self-reflection.
\newblock \emph{arXiv preprint arXiv:2310.11511}.

\bibitem[{Bang et~al.(2023)Bang, Cahyawijaya, Lee, Dai, Su, Wilie, Lovenia, Ji, Yu, Chung et~al.}]{bang2023multitask}
Yejin Bang, Samuel Cahyawijaya, Nayeon Lee, Wenliang Dai, Dan Su, Bryan Wilie, Holy Lovenia, Ziwei Ji, Tiezheng Yu, Willy Chung, et~al. 2023.
\newblock A multitask, multilingual, multimodal evaluation of chatgpt on reasoning, hallucination, and interactivity.
\newblock \emph{arXiv preprint arXiv:2302.04023}.

\bibitem[{Cormack et~al.(2009)Cormack, Clarke, and Buettcher}]{cormack2009reciprocal}
Gordon~V Cormack, Charles~LA Clarke, and Stefan Buettcher. 2009.
\newblock Reciprocal rank fusion outperforms condorcet and individual rank learning methods.
\newblock In \emph{Proceedings of the 32nd international ACM SIGIR conference on Research and development in information retrieval}, pages 758--759.

\bibitem[{Fränti and Mariescu-Istodor(2023)}]{FRANTI2023115}
Pasi Fränti and Radu Mariescu-Istodor. 2023.
\newblock \href {https://doi.org/10.1016/j.patrec.2023.02.005} {Soft precision and recall}.
\newblock \emph{Pattern Recognition Letters}, 167:115--121.

\bibitem[{Google(2024)}]{gemini-deep-research}
Google. 2024.
\newblock Try deep research and our new experimental model in gemini, your {AI} assistant.
\newblock \url{https://blog.google/products/gemini/google-gemini-deep-research/}.
\newblock Accessed: February 10, 2025.

\bibitem[{Guo et~al.(2025)Guo, Yang, Zhang, Song, Zhang, Xu, Zhu, Ma, Wang, Bi et~al.}]{guo2025deepseek}
Daya Guo, Dejian Yang, Haowei Zhang, Junxiao Song, Ruoyu Zhang, Runxin Xu, Qihao Zhu, Shirong Ma, Peiyi Wang, Xiao Bi, et~al. 2025.
\newblock Deepseek-r1: Incentivizing reasoning capability in llms via reinforcement learning.
\newblock \emph{arXiv preprint arXiv:2501.12948}.

\bibitem[{J{\"a}rvelin and Kek{\"a}l{\"a}inen(2002)}]{jarvelin2002cumulated}
Kalervo J{\"a}rvelin and Jaana Kek{\"a}l{\"a}inen. 2002.
\newblock Cumulated gain-based evaluation of ir techniques.
\newblock \emph{ACM Transactions on Information Systems (TOIS)}, 20(4):422--446.

\bibitem[{Johnson et~al.(2019)Johnson, Douze, and J{\'e}gou}]{johnson2019billion}
Jeff Johnson, Matthijs Douze, and Herv{\'e} J{\'e}gou. 2019.
\newblock Billion-scale similarity search with {GPUs}.
\newblock \emph{IEEE Transactions on Big Data}, 7(3):535--547.

\bibitem[{Khot et~al.(2022)Khot, Trivedi, Finlayson, Fu, Richardson, Clark, and Sabharwal}]{khot2022decomposed}
Tushar Khot, Harsh Trivedi, Matthew Finlayson, Yao Fu, Kyle Richardson, Peter Clark, and Ashish Sabharwal. 2022.
\newblock Decomposed prompting: A modular approach for solving complex tasks.
\newblock \emph{arXiv preprint arXiv:2210.02406}.

\bibitem[{Laskar et~al.(2023)Laskar, Bari, Rahman, Bhuiyan, Joty, and Huang}]{laskar2023systematic}
Md~Tahmid~Rahman Laskar, M~Saiful Bari, Mizanur Rahman, Md~Amran~Hossen Bhuiyan, Shafiq Joty, and Jimmy~Xiangji Huang. 2023.
\newblock A systematic study and comprehensive evaluation of chatgpt on benchmark datasets.
\newblock \emph{arXiv preprint arXiv:2305.18486}.

\bibitem[{Li et~al.(2021)Li, Fabbri, Kawamura, Liu, Tang, Tae, Shen, Ma, Mizutani, and Radev}]{li2021surfer100}
Irene Li, Alexander Fabbri, Rina Kawamura, Yixin Liu, Xiangru Tang, Jaesung Tae, Chang Shen, Sally Ma, Tomoe Mizutani, and Dragomir Radev. 2021.
\newblock Surfer100: Generating surveys from web resources, wikipedia-style.
\newblock \emph{arXiv preprint arXiv:2112.06377}.

\bibitem[{Liang et~al.(2022)Liang, Bommasani, Lee, Tsipras, Soylu, Yasunaga, Zhang, Narayanan, Wu, Kumar et~al.}]{liang2022holistic}
Percy Liang, Rishi Bommasani, Tony Lee, Dimitris Tsipras, Dilara Soylu, Michihiro Yasunaga, Yian Zhang, Deepak Narayanan, Yuhuai Wu, Ananya Kumar, et~al. 2022.
\newblock Holistic evaluation of language models.
\newblock \emph{arXiv preprint arXiv:2211.09110}.

\bibitem[{Liu et~al.(2018)Liu, Saleh, Pot, Goodrich, Sepassi, Kaiser, and Shazeer}]{liu2018generating}
Peter~J Liu, Mohammad Saleh, Etienne Pot, Ben Goodrich, Ryan Sepassi, Lukasz Kaiser, and Noam Shazeer. 2018.
\newblock Generating wikipedia by summarizing long sequences.
\newblock \emph{arXiv preprint arXiv:1801.10198}.

\bibitem[{Liu et~al.(2023{\natexlab{a}})Liu, Cao, Yang, and Wen}]{liu2023generating}
Shuaiqi Liu, Jiannong Cao, Ruosong Yang, and Zhiyuan Wen. 2023{\natexlab{a}}.
\newblock Generating a structured summary of numerous academic papers: Dataset and method.
\newblock \emph{arXiv preprint arXiv:2302.04580}.

\bibitem[{Liu et~al.(2023{\natexlab{b}})Liu, Yu, Zhang, Xu, Lei, Lai, Gu, Ding, Men, Yang et~al.}]{liu2023agentbench}
Xiao Liu, Hao Yu, Hanchen Zhang, Yifan Xu, Xuanyu Lei, Hanyu Lai, Yu~Gu, Hangliang Ding, Kaiwen Men, Kejuan Yang, et~al. 2023{\natexlab{b}}.
\newblock Agentbench: Evaluating llms as agents.
\newblock \emph{arXiv preprint arXiv:2308.03688}.

\bibitem[{Lo et~al.(2019)Lo, Wang, Neumann, Kinney, and Weld}]{lo2019s2orc}
Kyle Lo, Lucy~Lu Wang, Mark Neumann, Rodney Kinney, and Dan~S Weld. 2019.
\newblock S2orc: The semantic scholar open research corpus.
\newblock \emph{arXiv preprint arXiv:1911.02782}.

\bibitem[{Mihalcea and Tarau(2004)}]{mihalcea2004textrank}
Rada Mihalcea and Paul Tarau. 2004.
\newblock Textrank: Bringing order into text.
\newblock In \emph{Proceedings of the 2004 conference on empirical methods in natural language processing}, pages 404--411.

\bibitem[{OpenAI(2025)}]{openai-deep-research}
OpenAI. 2025.
\newblock Introducing deep research.
\newblock \url{https://openai.com/index/introducing-deep-research/}.
\newblock Accessed: February 10, 2025.

\bibitem[{Qian et~al.(2023)Qian, Cong, Yang, Chen, Su, Xu, Liu, and Sun}]{qian2023communicative}
Chen Qian, Xin Cong, Cheng Yang, Weize Chen, Yusheng Su, Juyuan Xu, Zhiyuan Liu, and Maosong Sun. 2023.
\newblock Communicative agents for software development.
\newblock \emph{arXiv preprint arXiv:2307.07924}.

\bibitem[{Qin et~al.(2023{\natexlab{a}})Qin, Zhang, Zhang, Chen, Yasunaga, and Yang}]{qin2023chatgpt}
Chengwei Qin, Aston Zhang, Zhuosheng Zhang, Jiaao Chen, Michihiro Yasunaga, and Diyi Yang. 2023{\natexlab{a}}.
\newblock Is chatgpt a general-purpose natural language processing task solver?
\newblock \emph{arXiv preprint arXiv:2302.06476}.

\bibitem[{Qin et~al.(2023{\natexlab{b}})Qin, Liang, Ye, Zhu, Yan, Lu, Lin, Cong, Tang, Qian et~al.}]{qin2023toolllm}
Yujia Qin, Shihao Liang, Yining Ye, Kunlun Zhu, Lan Yan, Yaxi Lu, Yankai Lin, Xin Cong, Xiangru Tang, Bill Qian, et~al. 2023{\natexlab{b}}.
\newblock Toolllm: Facilitating large language models to master 16000+ real-world apis.
\newblock \emph{arXiv preprint arXiv:2307.16789}.

\bibitem[{Reimers and Gurevych(2019)}]{reimers-gurevych-2019-sentence}
Nils Reimers and Iryna Gurevych. 2019.
\newblock \href {https://doi.org/10.18653/v1/D19-1410} {Sentence-{BERT}: Sentence embeddings using {S}iamese {BERT}-networks}.
\newblock In \emph{Proceedings of the 2019 Conference on Empirical Methods in Natural Language Processing and the 9th International Joint Conference on Natural Language Processing (EMNLP-IJCNLP)}, pages 3982--3992, Hong Kong, China. Association for Computational Linguistics.

\bibitem[{Rohman(1965)}]{rohman1965pre}
D~Gordon Rohman. 1965.
\newblock Pre-writing: The stage of discovery in the writing process.
\newblock \emph{College Composition \& Communication}, 16(2):106--112.

\bibitem[{Rosset et~al.(2024)Rosset, Chung, Qin, Chau, Feng, Awadallah, Neville, and Rao}]{rosset2024researchy}
Corby Rosset, Ho-Lam Chung, Guanghui Qin, Ethan~C Chau, Zhuo Feng, Ahmed Awadallah, Jennifer Neville, and Nikhil Rao. 2024.
\newblock Researchy questions: A dataset of multi-perspective, decompositional questions for llm web agents.
\newblock \emph{arXiv preprint arXiv:2402.17896}.

\bibitem[{Shao et~al.(2024)Shao, Jiang, Kanell, Xu, Khattab, and Lam}]{shao2024assisting}
Yijia Shao, Yucheng Jiang, Theodore~A Kanell, Peter Xu, Omar Khattab, and Monica~S Lam. 2024.
\newblock Assisting in writing wikipedia-like articles from scratch with large language models.
\newblock \emph{arXiv preprint arXiv:2402.14207}.

\bibitem[{Valenzuela et~al.(2015)Valenzuela, Ha, and Etzioni}]{valenzuela2015identifying}
Marco Valenzuela, Vu~Ha, and Oren Etzioni. 2015.
\newblock Identifying meaningful citations.
\newblock In \emph{Workshops at the twenty-ninth AAAI conference on artificial intelligence}.

\bibitem[{Voorhees(1999)}]{voorhees1999proceedings}
EM~Voorhees. 1999.
\newblock Proceedings of the 8th text retrieval conference.
\newblock \emph{TREC-8 Question Answering Track Report}, pages 77--82.

\bibitem[{Wang et~al.(2024)Wang, Li, Song, Xu, Tang, Zhuge, Pan, Song, Li, Singh, Tran, Li, Ma, Zheng, Qian, Shao, Muennighoff, Zhang, Hui, Lin, Brennan, Peng, Ji, and Neubig}]{openhands}
Xingyao Wang, Boxuan Li, Yufan Song, Frank~F. Xu, Xiangru Tang, Mingchen Zhuge, Jiayi Pan, Yueqi Song, Bowen Li, Jaskirat Singh, Hoang~H. Tran, Fuqiang Li, Ren Ma, Mingzhang Zheng, Bill Qian, Yanjun Shao, Niklas Muennighoff, Yizhe Zhang, Binyuan Hui, Junyang Lin, Robert Brennan, Hao Peng, Heng Ji, and Graham Neubig. 2024.
\newblock \href {https://arxiv.org/abs/2407.16741} {{OpenHands: An Open Platform for AI Software Developers as Generalist Agents}}.
\newblock \emph{Preprint}, arXiv:2407.16741.

\bibitem[{Xiao et~al.(2023)Xiao, Liu, Zhang, and Muennighof}]{xiao2023c}
Shitao Xiao, Zheng Liu, Peitian Zhang, and Niklas Muennighof. 2023.
\newblock C-pack: Packaged resources to advance general chinese embedding.
\newblock \emph{arXiv preprint arXiv:2309.07597}.

\bibitem[{Zhang and Shasha(1989)}]{zhang1989simple}
Kaizhong Zhang and Dennis Shasha. 1989.
\newblock Simple fast algorithms for the editing distance between trees and related problems.
\newblock \emph{SIAM journal on computing}, 18(6):1245--1262.

\bibitem[{Zhou et~al.(2023)Zhou, Xu, Zhu, Zhou, Lo, Sridhar, Cheng, Bisk, Fried, Alon et~al.}]{zhou2023webarena}
Shuyan Zhou, Frank~F Xu, Hao Zhu, Xuhui Zhou, Robert Lo, Abishek Sridhar, Xianyi Cheng, Yonatan Bisk, Daniel Fried, Uri Alon, et~al. 2023.
\newblock Webarena: A realistic web environment for building autonomous agents.
\newblock \emph{arXiv preprint arXiv:2307.13854}.

\end{thebibliography}
    \appendix

\section{Prompts for the Dataset Collection} \label{appendix:colleciton-prompts}

Instructions to analyze whether an academic paper fits under the research survey category.

\begin{lstlisting}[
    basicstyle=\ttfamily\small, % Adjust the font size as needed
    breaklines=true,            % Enable automatic line breaking
    breakatwhitespace=true,     % Break lines at whitespace
    columns=flexible,           % Improve alignment of wrapped lines
    frame=single,               % Optional: Adds a box around the code
]
"""
The point of a survey paper is to provide an organized view on the current state of the field. If it relies heavily on external information, such as the results of a population questionnaire, do not include it. Using the above criteria, is the following article a survey paper? Respond either "True" or "False".
"""
\end{lstlisting}

Instructions to extract the survey mind-maps into JSON-encoded representations.

\begin{lstlisting}[
    basicstyle=\ttfamily\small, % Adjust the font size as needed
    breaklines=true,            % Enable automatic line breaking
    breakatwhitespace=true,     % Break lines at whitespace
    columns=flexible,           % Improve alignment of wrapped lines
    frame=single,               % Optional: Adds a box around the code
]
"""
Identify the figure that most likely illustrates a taxonomy or overview. Your response should be limited to the filename, or NULL if not found. The provided figure presents a hierarchy. Extract as JSON-encoded tree whose children are NULL-terminated.
"""
\end{lstlisting}

\section{Prompts for the Experiments} \label{appendix:experiment-prompts}

Instructions used by \textsc{Decomposer} for the information discovery task, adopted from the Researchy Questions by \citet{rosset2024researchy}.

\begin{lstlisting}[
    basicstyle=\ttfamily\small, % Adjust the font size as needed
    breaklines=true,            % Enable automatic line breaking
    breakatwhitespace=true,     % Break lines at whitespace
    columns=flexible,           % Improve alignment of wrapped lines
    frame=single,               % Optional: Adds a box around the code
]
"""
### Below is an example on how to decompose a complex question into sub-questions and search queries.

Question: should the death penalty be legalized?

<Decomposition>
    - What are the arguments in favor of the death penalty?
        - Does the death penalty serve as a deterrent to crime?
        - Is the death penalty a just punishment for certain crimes?
        - How does the death penalty compare to other forms of punishment in terms of cost and effectiveness?
    - What are the arguments against the death penalty?
        - What is the risk of executing innocent people with a death penalty?
        - Are there any ethical concerns surrounding the death penalty?
        - To what extent is the death penalty applied fairly and without bias?
        - In practice, how expensive is the death penalty?
    - What is the current legal status of the death penalty in various jurisdictions?
        - In which countries or states is the death penalty currently legal?
        - What are the trends in death penalty legislation and public opinion?
    - What are the alternatives to the death penalty?
        - How effective are alternative punishments to the death penalty, e.g. life imprisonment?
        - What are the costs and benefits of alternatives to the death penalty?
    - How do the pros and cons of the death penalty compare to its alternatives?
</Decomposition>

<Queries>
    - arguments in favor of the death penalty
    - death penalty as a deterrent to crime
    - death penalty as a just punishment
    - death penalty cost and effectiveness comparison
    - arguments against the death penalty
    - risk of executing innocent people with death penalty
    - ethical concerns surrounding the death penalty
    - fairness and bias in death penalty application
    - current legal status of the death penalty worldwide
    - trends in death penalty legislation and public opinion
    - alternatives to the death penalty
    - effectiveness of life imprisonment without parole
    - costs and benefits of death penalty alternatives
</Queries>

Question: {x}

### Instructions:

1. What sub-questions do I need to know in order to fully understand and answer the above Question.
    - Format your response as a bullet-point style outline of questions and sub-questions in the <Decomposition> tag.
    - Order your sub-questions such that one question comes after another if it needs to use the answer to the previous one.
    - Do not ask unnecessary or tangential sub-questions, only those that are critical to finding important information.  
2) Next, write a list of search queries that would likely lead to results addressing all the sub-questions.
    - Enumerate your queries in a bullet-point style list inside the <Queries> tag.

You may refer to the example above for guidance.
"""
\end{lstlisting}

Instructions used by \textsc{Zero-Shot} and \textsc{Self-RAG} for the information discovery task.

\begin{lstlisting}[
    basicstyle=\ttfamily\small, % Adjust the font size as needed
    breaklines=true,            % Enable automatic line breaking
    breakatwhitespace=true,     % Break lines at whitespace
    columns=flexible,           % Improve alignment of wrapped lines
    frame=single,               % Optional: Adds a box around the code
]
"""
Create a search query that gathers supporting materials for writing a survey paper on the following topic: {x}.
"""
\end{lstlisting}

Instructions used by \textsc{Few-Shot} for the information organization task.

\begin{lstlisting}[
    basicstyle=\ttfamily\small, % Adjust the font size as needed
    breaklines=true,            % Enable automatic line breaking
    breakatwhitespace=true,     % Break lines at whitespace
    columns=flexible,           % Improve alignment of wrapped lines
    frame=single,               % Optional: Adds a box around the code
]
"""
### Examples

<topic>
A Survey on LiDAR Scanning Mechanisms
</topic>

<typology>
{"Opto-Mechanical Beam Deflection Mechanisms": {"Line Scanner": {"Slanted Plain Mirror": null, "Off-axis Parabolic Mirror": null, "Polygon Mirror": null}, "Area Scanner": {"Single Galvanometer Scanning Mirror": null, "Double Galvanometer Scanning Mirror": null, "Gyroscopic Mirror": null, "Risley Scanner": null}}}
</typology>

<topic>
A Survey on Large Language Models for Recommendation
</topic>

<typology>
{"LLM4Rec": {"Discriminative LLM4Rec": {"Fine-tuning": {"Prompt Tuning": null}}, "Generative LLM4Rec": {"Non-tuning": {"Prompting": null, "In-context Learning": null}, "Tuning": {"Fine-tuning": null, "Prompt Tuning": null, "Instruction Tuning": null}}}}
</typology>

### Instructions

- Provided a topic, your task is to construct a mind-map style typology that presents a systematic understanding of the topic. 
- Put your JSON-encoded response in the tag `<typology>...</typology>`. You may refer to the examples above for guidance.

<topic>
{x}
</topic>
"""
\end{lstlisting}

\section{Inspection over Survey Passages} \label{appendix:inspection-over-survey-passages}

A manual inspection of survey passages is conducted to assess the effectiveness of GPT-4 in identifying survey papers. Following this, the extracted mind-map is analyzed and compared against the original paper to determine its accuracy and relevance in capturing key aspects of the topic under discussion. Details are presented in Table \ref{table:inspection:1} and \ref{table:inspection:2}.

\section{Indexing Schemes} \label{appendix:indexing-schemes}

We present a comparative analysis of different document indexing schemes—title-only, abstract, and full-text—used during the information discovery stage. While the main experiments adopt abstract indexing for its balance between informativeness and tractability, we evaluated how alternative schemes affect retrieval performance across three baselines: \textsc{Title}, \textsc{Zero-Shot}, and \textsc{Decomposer} using \textsc{GPT-4}. Results are reported for both BM25 and BGE retrieval models in Table~\ref{table:indexing-schemes}.

\begin{table*}
    \small
    \caption{Impact of indexing schemes on retrieval performance.}
    \label{table:indexing-schemes}
    \centering
    \begin{tabular}{rcccccccc}
        \toprule
        & \multicolumn{2}{c}{Recall@10} & \multicolumn{2}{c}{Recall@100} & \multicolumn{2}{c}{Precision@10} & \multicolumn{2}{c}{Precision@100} \\
        \cmidrule(lr){2-3} \cmidrule(lr){4-5} \cmidrule(lr){6-7} \cmidrule(lr){8-9}
        Method (Indexing) & BM25 & BGE & BM25 & BGE & BM25 & BGE & BM25 & BGE \\
        \midrule
        \textsc{Title} (Title) & 0.0150 & \textbf{0.0184} & 0.0347 & \textbf{0.0419} & 0.0297 & \textbf{0.0378} & 0.0082 & \textbf{0.0101} \\
        \textsc{Title} (Full-Text) & 0.0038 & 0.0018 & 0.0072 & 0.0027 & 0.0091 & 0.0054 & 0.0020 & 0.0009 \\
        \textsc{Zero-Shot} (Title) & 0.0122 & 0.0122 & 0.0292 & 0.0293 & 0.0245 & 0.0238 & 0.0069 & 0.0067 \\
        \textsc{Zero-Shot} (Full-Text) & 0.0035 & 0.0016 & 0.0067 & 0.0024 & 0.0081 & 0.0046 & 0.0019 & 0.0008 \\
        \textsc{Decomposer} (Title) & 0.0125 & 0.0130 & 0.0297 & 0.0305 & 0.0238 & 0.0250 & 0.0070 & 0.0072 \\
        \textsc{Decomposer} (Full-Text) & 0.0037 & 0.0017 & 0.0076 & 0.0026 & 0.0088 & 0.0049 & 0.0022 & 0.0008 \\
        \bottomrule
    \end{tabular}
\end{table*}

\section{Case Study with Gemini Deep Research} \label{appendix:case-study-with-gemini-deep-research}

\begin{figure*}
    \centering
    \includegraphics[width=\linewidth]{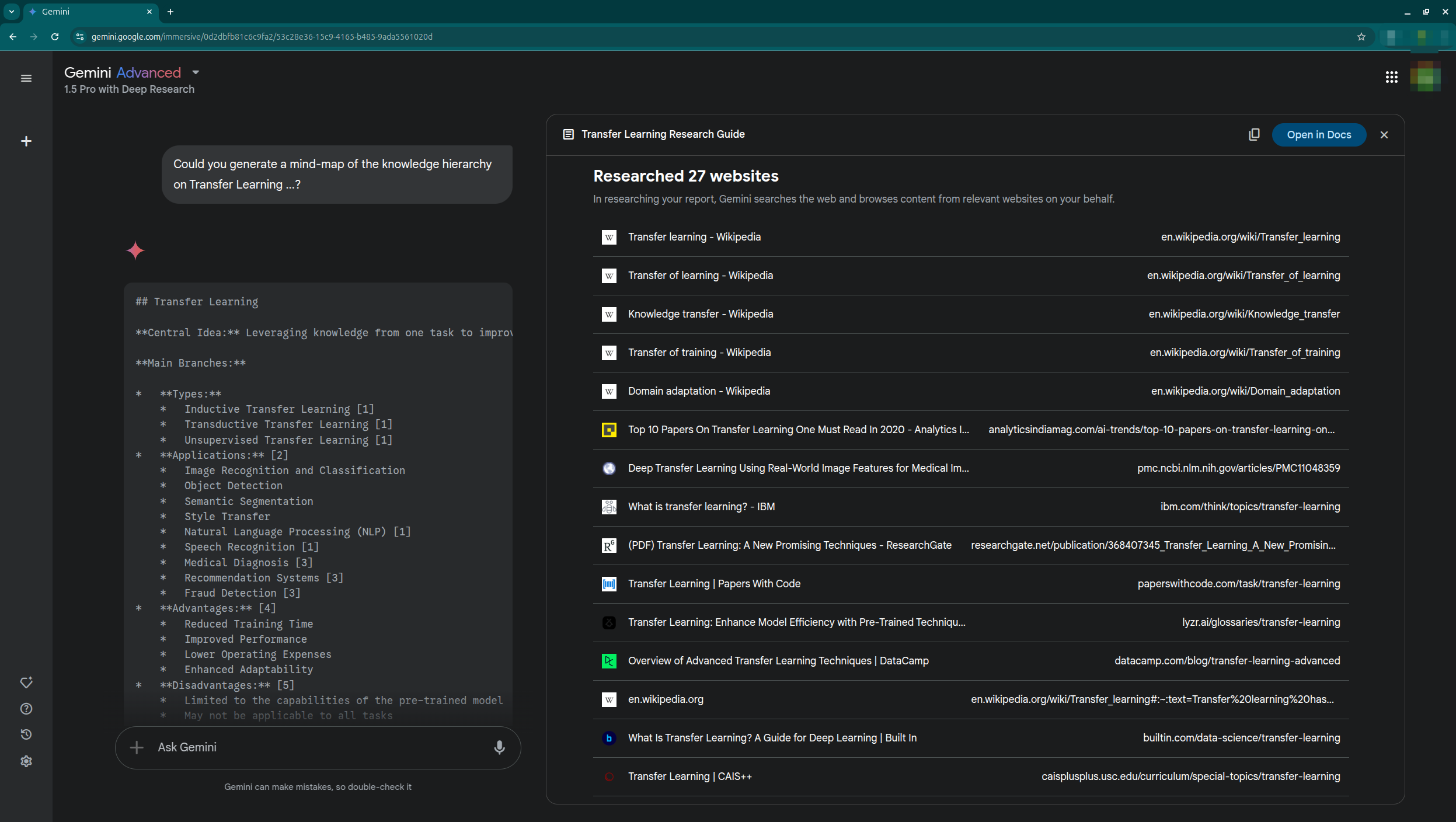}
    \caption{\label{figure:deepresearch-2} Sources used while asking Deep Research from Gemini to work on transfer learning.}
\end{figure*}

Generated report on the topic of LiDAR scanning mechanisms with sources attached in Figure \ref{figure:deepresearch-2}.

\begin{lstlisting}[
    basicstyle=\ttfamily\small, % Adjust the font size as needed
    breaklines=true,            % Enable automatic line breaking
    breakatwhitespace=true,     % Break lines at whitespace
    columns=flexible,           % Improve alignment of wrapped lines
    frame=single,               % Optional: Adds a box around the code
]
# **LiDAR Scanning Mechanisms: A Comprehensive Overview**

LiDAR (Light Detection and Ranging) technology has revolutionized various industries, from autonomous vehicles and robotics to surveying and mapping. At the heart of LiDAR systems lie the scanning mechanisms that enable these systems to capture precise 3D spatial information about the surrounding environment. This article delves into the different types of LiDAR scanning mechanisms, exploring their working principles, advantages, disadvantages, applications, and the companies that manufacture them. We will also examine the latest advancements and future trends in LiDAR scanning technology.

## **Types of LiDAR Scanning Mechanisms**

LiDAR scanning mechanisms can be broadly classified into three main categories:

* **Mechanical:** These systems utilize moving parts, such as rotating mirrors, to steer the laser beam and scan the environment1.  
* **Solid-state:** These systems have no macroscopic moving parts and employ techniques like MEMS mirrors or optical phased arrays to steer the laser beam electronically1.  
* **Flash:** These systems illuminate the entire scene with a single laser pulse, capturing a 3D point cloud in an instant, similar to a camera1.

## **Mechanical LiDAR**

Mechanical LiDAR, the oldest type of LiDAR, employs a rotating assembly with mirrors or prisms to direct the laser beam in a 360-degree scan1. This mechanism allows for a wide field of view and high resolution, making it suitable for applications that require detailed 3D mapping2.

### **How it Works**

A mechanical LiDAR system typically consists of a laser emitter, a rotating mirror or prism assembly, and a detector. The laser emits short pulses of light, which are directed by the rotating assembly towards the target area. The detector measures the time it takes for the light to return, allowing the system to calculate the distance to the object4. The two kinds of LiDAR detection schemes are "incoherent" or direct energy detection (which principally measures amplitude changes of the reflected light) and coherent detection (best for measuring Doppler shifts, or changes in the phase of the reflected light)3. Coherent systems generally use optical heterodyne detection. This is more sensitive than direct detection and allows them to operate at much lower power, but requires more complex transceivers. Both types employ pulse models: either micropulse or high energy. Micropulse systems utilize intermittent bursts of energy.

### **Advantages**

* **Wide Field of View:** Mechanical LiDAR can achieve a 360-degree horizontal field of view, providing comprehensive coverage of the surrounding environment5.  
* **High Resolution:** The rotating mechanism allows for precise control over the laser beam, enabling high-resolution data capture2.  
* **Long Range:** Mechanical LiDAR systems can achieve long-range measurements, making them suitable for applications like aerial surveying and mapping2.

### **Disadvantages**

* **Moving Parts:** The presence of moving parts makes mechanical LiDAR systems susceptible to wear and tear, potentially affecting their reliability and lifespan2.  
* **Bulkiness:** Mechanical systems tend to be larger and heavier than solid-state LiDAR, making them less suitable for applications where size and weight are critical5.  
* **Cost:** The complexity of the mechanical design can contribute to higher manufacturing costs6.  
* **Environmental Factors:** LiDAR signals can be attenuated or scattered by fog, rain, and snow, which can limit their effectiveness in adverse weather conditions7. Additionally, LiDAR technology faces challenges in detecting and measuring certain types of surfaces, particularly non-reflective or highly absorbing ones. For instance, black asphalt, dark-colored objects, or water surfaces may not reflect sufficient energy from the laser pulses, making them difficult to detect or measure accurately7.

## **Solid-State LiDAR**

Solid-state LiDAR eliminates the need for macroscopic moving parts, offering improved durability and reliability compared to mechanical systems2. These systems utilize various technologies to steer the laser beam electronically, including MEMS (Microelectromechanical Systems) and OPA (Optical Phased Array).

### **How it Works**

#### **MEMS LiDAR**

MEMS LiDAR works by directing a single laser beam to a tiny mirror that can be tilted or rotated to scan the environment3. The size of the MEMS mirror is a critical factor in its performance8. A larger mirror allows for more photons to be emitted, increasing the chances of a sufficient number of photons returning to the detector for object detection. However, the mirror must also be large enough to deflect all the light collimated by the lens used to focus the laser beam. This ensures high resolution and accurate identification of even small objects8. While some MEMS systems operate in a single plane, others can achieve 2D scanning with dual-axis mirrors or multiple lasers3.

#### **OPA LiDAR**

OPA LiDAR, on the other hand, uses an array of optical antennas to create a beam that can be steered electronically by controlling the phase of the light emitted from each antenna1.

### **MEMS vs. Polygon Scanners**

MEMS mirrors and polygon scanners are both used in LiDAR systems for beam steering, but they have distinct characteristics and trade-offs9. MEMS mirrors are smaller and potentially more cost-effective, but they can be more susceptible to vibrations and temperature variations. Polygon scanners, with their larger size and more robust design, offer higher accuracy and longer range, but they can be more expensive and less compact.

### **Advantages**

* **Durability:** The absence of macroscopic moving parts makes solid-state LiDAR more robust and less prone to mechanical failure2.  
* **Compact Size:** Solid-state LiDAR systems are typically smaller and lighter than mechanical systems, making them suitable for applications where space is limited6.  
* **Faster Scanning:** Electronic beam steering allows for faster scanning speeds compared to mechanical systems10.

### **Disadvantages**

* **Limited Field of View:** Solid-state LiDAR may have a more limited field of view compared to mechanical systems, although advancements in beam steering technology are addressing this limitation5.  
* **Shorter Range:** Solid-state LiDAR typically has a shorter range than mechanical LiDAR, although this is also improving with advancements in technology5.  
* **Cost:** While the cost of solid-state LiDAR is decreasing, it can still be higher than some mechanical systems11.

## **Flash LiDAR**

Flash LiDAR illuminates the entire scene with a single laser pulse, capturing a 3D point cloud instantaneously1. This approach eliminates the need for scanning mechanisms, resulting in a simpler and potentially more cost-effective design.

### **How it Works**

Flash LiDAR systems use a wide-beam laser to illuminate the entire field of view. The reflected light is then captured by a detector array, which measures the time of flight for each pixel in the array. This allows the system to generate a 3D image of the scene in a single flash4.

### **Advantages**

* **No Moving Parts:** Flash LiDAR has no moving parts, making it highly durable and reliable1.  
* **Fast Data Acquisition:** It captures an entire scene in a single flash, enabling rapid data acquisition2.  
* **Compact Size:** Flash LiDAR systems can be very compact, making them suitable for integration into small devices1.

### **Disadvantages**

* **Limited Range:** Flash LiDAR typically has a shorter range compared to scanning LiDAR systems1.  
* **Lower Resolution:** The resolution of flash LiDAR can be lower than scanning LiDAR, especially at longer distances1.  
* **Eye Safety:** The high-power laser pulses used in flash LiDAR can pose eye safety concerns, requiring careful design and implementation7. The use of eye-safe wavelengths, such as 1550 nm, is one approach to mitigate this concern3.

## **Hybrid LiDAR**

Hybrid LiDAR systems combine elements of both solid-state and mechanical LiDAR technologies to optimize performance and address the limitations of each approach2.

### **How it Works**

Hybrid LiDAR systems may use a combination of MEMS mirrors and rotating elements to achieve a wider field of view and longer range. For example, the Hesai Pandar128, a hybrid solid-state LiDAR, integrates 128 transmit-receive modules with a 360-degree spinning scanning module12. This combination allows for high resolution and a wide field of view while maintaining a compact design.

### **Advantages**

* **Balanced Performance:** Hybrid LiDAR offers a balance between the accuracy and range of mechanical systems and the durability and compact design of solid-state systems2.  
* **Adaptability:** It can adapt to the needs of different environments by leveraging the reliability of solid-state components and the detailed scanning capabilities of mechanical systems2.

### **Disadvantages**

* **Complexity:** Hybrid systems can be more complex to design and manufacture compared to purely mechanical or solid-state systems13.  
* **Cost:** The combination of technologies can lead to higher costs compared to some single-technology systems13.

## **Cost and Availability of LiDAR Systems**

The cost and availability of LiDAR systems vary significantly depending on several factors, including performance specifications, application requirements, durability, and integration needs14.

* **Performance Specifications:** Higher performance systems with greater range, resolution, and speed are generally more expensive14. For example, automotive-grade LiDAR, with its long-range and high-resolution requirements, can be significantly more costly than LiDAR used for general surveying14.  
* **Application Requirements:** LiDAR systems designed for demanding applications, such as autonomous vehicles, are typically more expensive due to their stringent performance and reliability requirements14.  
* **Durability and Reliability:** LiDAR systems built for harsh environments require robust designs and materials, which can increase costs14.  
* **Integration and Maintenance:** Systems that are easier to integrate and maintain can be more cost-effective in the long run14.

The cost of LiDAR systems has been decreasing in recent years due to advancements in technology and manufacturing15. However, high-performance systems can still be expensive. For example, a robust, entry-level LiDAR system for drone applications can cost around $23,000, while a drone and associated accessories can add another $10,000 to $26,000 to the total cost16.

## **LiDAR vs. Radar**

While both LiDAR and radar are remote sensing technologies used for object detection and mapping, they have distinct characteristics and strengths7. LiDAR uses laser light to measure distances and create high-resolution 3D maps, while radar uses radio waves to detect objects and measure their speed and direction. LiDAR offers higher accuracy and resolution, especially for detailed mapping and object recognition, but it can be more sensitive to adverse weather conditions. Radar, on the other hand, is less affected by weather and can operate at longer ranges, but it typically provides lower resolution data.

## **Applications of LiDAR**

LiDAR technology has a wide range of applications across various industries, including:

* **Autonomous Vehicles:** LiDAR is a crucial sensor in self-driving cars, enabling them to perceive their surroundings, navigate autonomously, and avoid obstacles17.  
* **Robotics:** LiDAR helps robots navigate, map their environment, and interact with objects17.  
* **Surveying and Mapping:** LiDAR is used to create detailed 3D maps of terrain, infrastructure, and urban environments17.  
* **Environmental Monitoring:** LiDAR is used to monitor vegetation, assess natural disaster impacts, and map forests17.  
* **Urban Planning and Development:** LiDAR provides data for urban planning, infrastructure development, and transportation management17.  
* **Archaeology:** LiDAR helps uncover ancient structures and features hidden beneath vegetation or over time17.  
* **Disaster Management:** LiDAR is used to assess damage after natural disasters and aid in rescue efforts17.

## **Latest Advancements in LiDAR Scanning Mechanisms**

Research and development in LiDAR technology continue to advance, leading to innovative solutions that improve performance, reduce costs, and expand applications. Some of the latest advancements include:

* **Elastic LiDAR:** This technology utilizes elastic scattering of light to achieve precise measurements and environmental sensing18.  
* **Silicon-based LiDAR:** This approach leverages silicon photonics to create compact and cost-effective LiDAR systems. Silicon-based LiDAR is an ideal way to reduce the volume of the LiDAR and realize monolithic integration. It removes the moving parts in the conventional device and realizes solid-state beam steering19.  
* **Advanced Simulation Tools:** Tools like VPItransmissionMaker and VPIcomponentMaker enable detailed design and simulation of LiDAR systems, including modeling atmospheric conditions and photonic integrated circuits20.  
* **Smart Corner Solution:** Marelli's Smart Corner solution integrates LiDAR sensors into vehicle headlamps and grilles, addressing sensor placement and field of view challenges20.  
* **Deep Learning for Drone Detection:** Recent research has demonstrated the effectiveness of deep learning algorithms in processing LiDAR data for drone detection21. Studies have shown high accuracy (above 97%) in detecting drones even in noisy environments using deep learning models21.

## **The Future of LiDAR Scanning Mechanisms**

The future of LiDAR scanning mechanisms is likely to be shaped by several key trends:

* **Continued Miniaturization:** Solid-state and flash LiDAR technologies will continue to shrink in size, enabling their integration into smaller devices and expanding their applications15.  
* **Improved Performance:** Advancements in laser technology, detectors, and scanning mechanisms will lead to higher resolution, longer range, and faster scanning speeds15.  
* **Reduced Costs:** Innovations in manufacturing and design will drive down the cost of LiDAR systems, making them more accessible for various applications15.  
* **Increased Integration:** LiDAR systems will be increasingly integrated with other sensors and technologies, such as cameras and radar, to create more comprehensive and robust sensing solutions22.

## **Conclusion**

LiDAR scanning mechanisms are fundamental to the operation and application of LiDAR technology. Each type of scanning mechanism, whether mechanical, solid-state, flash, or hybrid, offers a unique set of advantages and disadvantages. Mechanical systems excel in providing a wide field of view and high resolution, making them suitable for applications like aerial surveying and mapping. Solid-state LiDAR, with its MEMS and OPA variants, offers durability, compact size, and faster scanning speeds, making it increasingly popular for autonomous vehicles and robotics. Flash LiDAR, while limited in range and resolution, provides rapid data acquisition and is suitable for short-range applications. Hybrid systems combine the strengths of different approaches to optimize performance for specific needs.

The choice of LiDAR scanning mechanism depends on the specific application requirements, considering factors such as range, resolution, field of view, cost, and environmental conditions. As LiDAR technology continues to evolve, driven by advancements in miniaturization, performance improvement, and cost reduction, we can expect to see even more innovative scanning mechanisms emerge, further expanding the applications and impact of this transformative technology.
\end{lstlisting}

Generated report on the topic of transfer learning with sources attached in Figure \ref{figure:deepresearch-2}.

\begin{lstlisting}[
    basicstyle=\ttfamily\small, % Adjust the font size as needed
    breaklines=true,            % Enable automatic line breaking
    breakatwhitespace=true,     % Break lines at whitespace
    columns=flexible,           % Improve alignment of wrapped lines
    frame=single,               % Optional: Adds a box around the code
]
# **Transfer Learning: A Comprehensive Overview**

## **1\. Introduction**

Transfer learning, a powerful machine learning technique, has emerged as a game-changer in the field of artificial intelligence (AI). It leverages the knowledge gained from solving one problem to improve performance on a different but related problem. This approach has gained significant traction in recent years, particularly in deep learning, where training models from scratch can be computationally expensive and require massive amounts of data. By transferring knowledge from pre-trained models, transfer learning enables faster training, better performance, and the ability to tackle problems with limited data1.

## **2\. Background**

The concept of transfer learning is rooted in the idea that humans can apply knowledge learned in one context to new situations. For example, learning to ride a bicycle can make it easier to learn to ride a motorcycle. Similarly, in machine learning, transfer learning allows models to leverage pre-existing knowledge to accelerate learning and improve performance on new tasks2.

Early research on transfer learning dates back to the 1976s, with studies exploring knowledge transfer in neural networks1. Over the years, the field has evolved, with significant contributions from researchers like Lorien Pratt, who formulated the discriminability-based transfer (DBT) algorithm in 19921. By 1998, the field had expanded to include multi-task learning and more formal theoretical foundations1.

Andrew Ng, a prominent figure in AI, highlighted the importance of transfer learning in his NIPS 2016 tutorial, predicting that it would become a key driver of machine learning commercial success1. This prediction has come to fruition, with transfer learning now playing a crucial role in various AI applications, including image recognition, natural language processing, and speech recognition.

## **3\. Types of Transfer Learning**

Transfer learning can be categorized into different types based on the relationship between the source and target tasks and domains. Three common types are:

* **Inductive Transfer Learning:** In this type, the source and target tasks are different, but the domains are the same. This is often used in computer vision, where models pre-trained on large image datasets are adapted for specific tasks like object detection3.  
* **Transductive Transfer Learning:** Here, the source and target tasks are the same, but the domains are different. For example, a model trained on restaurant reviews could be adapted to classify movie reviews3.  
* **Unsupervised Transfer Learning:** This type involves unlabeled data in both the source and target domains. It is similar to inductive transfer learning but focuses on unsupervised tasks3.

These types of transfer learning offer flexibility in adapting models to different scenarios, depending on the availability of labeled data and the similarity between tasks and domains.

## **4\. Applications of Transfer Learning**

Transfer learning has found widespread applications in various domains, revolutionizing the way AI models are developed and deployed. Some notable applications include:

* **Image Recognition and Classification:** Transfer learning has significantly improved image recognition tasks by leveraging pre-trained models on large datasets like ImageNet. These models can be fine-tuned for specific tasks, such as medical image classification or identifying species in wildlife images4.  
* **Natural Language Processing (NLP):** Transfer learning has been instrumental in advancing NLP applications, including sentiment analysis, text classification, and machine translation. Pre-trained language models like BERT and GPT can be adapted for specific language processing tasks, enabling more accurate and efficient language understanding5.  
* **Speech Recognition:** Transfer learning has enhanced speech recognition systems by transferring knowledge from general audio models. This has led to improved accuracy in voice commands, transcription, and other speech-related tasks6.  
* **Medical Diagnosis:** Transfer learning has shown promise in improving medical diagnosis by adapting models trained on existing medical imaging datasets. This can aid in faster and more accurate diagnoses, leading to better patient outcomes7.  
* **Recommendation Systems:** Transfer learning can be used to improve recommendation systems by leveraging knowledge from user behavior data. This enables models to make more personalized recommendations and enhance user experiences6.

These are just a few examples of how transfer learning is being applied across different domains. Its ability to adapt models to new tasks and domains with limited data has made it a valuable tool in various AI applications.

## **5\. Advantages and Disadvantages of Transfer Learning**

Transfer learning offers several advantages over training models from scratch:

* **Reduced Training Time:** By leveraging pre-trained models, transfer learning significantly reduces the time required to train a model for a new task. This is because the model already has a foundation of knowledge, and only the final layers or specific parameters need to be adjusted8.  
* **Improved Performance:** Transfer learning often leads to better performance, especially when data for the new task is limited. The pre-trained model has already learned relevant features and patterns, which can be beneficial for the new task3.  
* **Lower Computational Costs:** Transfer learning can reduce computational costs by requiring less data and training time. This is particularly important in deep learning, where training models can be computationally expensive3.  
* **Enhanced Generalization:** Transfer learning can improve the generalization ability of models by incorporating knowledge from other domains. This helps models perform better on unseen data and reduces the risk of overfitting3.

However, transfer learning also has some limitations:

* **Domain Mismatch:** If the source and target domains are significantly different, transfer learning may not be effective. The pre-trained model may not have learned features relevant to the new task, leading to poor performance9.  
* **Overfitting:** Fine-tuning a pre-trained model on a small dataset can lead to overfitting, where the model performs well on the training data but poorly on unseen data10.  
* **Negative Transfer:** In some cases, transferring knowledge from the source domain can negatively impact the performance on the target task. This can happen if the tasks are dissimilar or the source domain has irrelevant features1.

Despite these limitations, the benefits of transfer learning often outweigh the drawbacks, making it a valuable technique in many machine learning applications.

## **6\. Future of Transfer Learning**

Transfer learning is an evolving field with ongoing research and development. Some key areas for future exploration include:

* **Multi-domain Adaptation:** Developing models that can effectively transfer knowledge across multiple diverse domains11.  
* **Incremental Learning:** Enabling models to continuously learn and adapt to new information while retaining previously learned knowledge12.  
* **Model Compression:** Reducing the size of large pre-trained models without sacrificing performance, making them more suitable for deployment in resource-constrained environments12.  
* **Addressing Ethical Concerns:** Ensuring fairness, mitigating bias, and addressing privacy concerns in transfer learning applications13.

These advancements will further enhance the capabilities of transfer learning and expand its applications in various fields.

## **7\. Conclusion**

Transfer learning has become a highly effective approach in machine learning, allowing for faster training, enhanced performance, and the ability to address challenges with limited data. By utilizing pre-trained models, it has transformed numerous AI applications, such as image recognition, natural language processing, and speech recognition. Despite certain challenges and limitations, continuous research and development are driving its evolution, ensuring an even greater impact on the future of AI.
\end{lstlisting}

\begin{figure*}
    \centering
    \includegraphics[width=\linewidth]{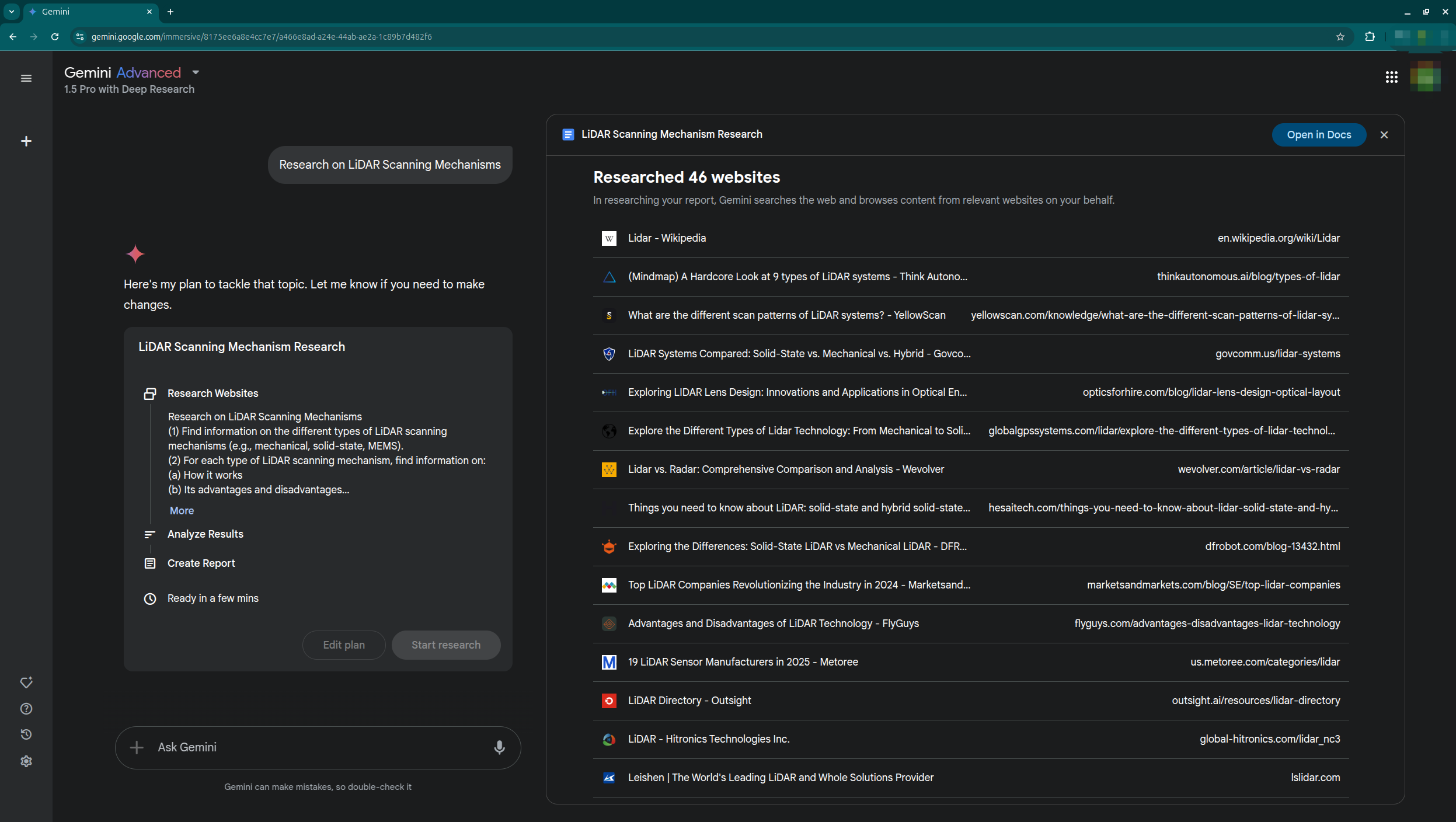}
    \caption{\label{figure:deepresearch-1} Sources used while asking Deep Research from Gemini to work on LiDAR Scanning Mechanisms.}
\end{figure*}

\begin{table}[t]
    \small
    \centering
    \caption{\label{table:inspection:1} Manual evaluation of survey selection and mind-map extraction for the first batch of 50 papers.}
    \begin{tabular}{cccc}
        \toprule
        CorpusID & IsSurvey & AccurateMap & RelevantMap\\
        \midrule
        3524264&True&True&True\\
        3644401&True&True&True\\
        4503761&True&True&False\\
        5058972&True&True&False\\
        8034133&True&True&True\\
        8922493&True&True&True\\
        9935621&True&False&True\\
        10934716&True&True&True\\
        17513321&True&True&True\\
        18750590&True&True&True\\
        20774863&False&False&False\\
        22727391&True&True&True\\
        23384543&True&True&True\\
        33413610&True&True&False\\
        52841074&True&True&True\\
        55254540&True&True&True\\
        56895323&True&True&True\\
        57721147&True&False&True\\
        64256250&True&False&True\\
        115523285&True&True&False\\
        119297355&True&False&True\\
        198933686&True&True&True\\
        210865204&True&True&False\\
        211010433&True&True&True\\
        212665971&True&True&True\\
        214743520&True&True&True\\
        216056393&True&True&True\\
        219316962&True&True&False\\
        220302470&True&False&True\\
        221446014&True&True&True\\
        222095837&True&True&False\\
        225029039&False&False&False\\
        226227376&True&True&False\\
        226300094&True&True&True\\
        227228021&True&True&False\\
        229363354&True&False&True\\
        231149960&False&False&False\\
        231698518&True&True&True\\
        233722066&False&False&False\\
        234213205&True&True&False\\
        235352671&True&False&False\\
        235458292&True&True&True\\
        235485414&True&True&False\\
        235490196&True&True&True\\
        235669589&True&True&True\\
        235766219&True&True&False\\
        236090307&True&True&True\\
        236772630&True&True&True\\
        236976256&True&True&True\\
        236986986&True&True&False\\
        \bottomrule
    \end{tabular}
\end{table}

\begin{table}[t]
    \small
    \centering
    \caption{\label{table:inspection:2} Manual evaluation of survey selection and mind-map extraction for the second batch of 50 papers.}
    \begin{tabular}{cccc}
        \toprule
        CorpusID & IsSurvey & AccurateMap & RelevantMap\\
        \midrule
        237372527&True&False&True\\
        237373628&True&True&True\\
        237485263&True&True&True\\
        238242214&True&False&True\\
        238242941&True&True&True\\
        238242941&True&False&False\\
        238639787&True&True&False\\
        244119139&True&True&True\\
        244773222&True&True&True\\
        245353469&True&True&True\\
        245877584&False&False&False\\
        247763152&True&True&True\\
        248834382&True&True&False\\
        249209981&True&True&False\\
        249687282&True&False&False\\
        250089226&True&True&True\\
        250939903&True&True&True\\
        251104722&True&True&True\\
        251506514&True&True&True\\
        251643467&True&True&False\\
        252683270&True&True&True\\
        252762319&True&True&False\\
        253370610&True&True&True\\
        253796900&True&True&True\\
        254274880&True&False&True\\
        254756520&True&True&True\\
        255025269&True&True&True\\
        256227178&True&True&True\\
        256826729&True&True&True\\
        257232619&True&True&False\\
        257255597&True&False&True\\
        257522478&True&True&False\\
        258539255&True&True&True\\
        258722762&True&True&True\\
        259043594&True&True&True\\
        259088696&True&True&True\\
        259108865&True&True&True\\
        259154569&False&False&False\\
        259283502&True&True&True\\
        259951356&True&True&True\\
        260229544&True&True&True\\
        260316174&True&True&True\\
        260849783&True&False&True\\
        261682162&True&False&True\\
        263134374&True&True&False\\
        263334211&True&True&True\\
        263830273&True&True&True\\
        263831409&True&False&True\\
        263909496&True&True&True\\
        264604532&True&True&True\\
        \bottomrule
    \end{tabular}
\end{table}

\end{document}